\documentclass[final,3p,times]{elsarticle} 

\usepackage[dvipsnames,svgnames,x11names]{xcolor}

\usepackage[draft]{changes}
\usepackage[inline]{enumitem}
\definechangesauthor[color=blue]{PC}
\usepackage{lineno,hyperref}
\usepackage{amsmath,amssymb,amsfonts}
\usepackage{algorithmicx}
\usepackage{booktabs}
\usepackage{graphicx}
\usepackage{caption}
\usepackage{subfigure}
\usepackage[font=small,labelfont=bf, figurename=Fig.]{caption} 

\usepackage{textcomp}
\usepackage{todonotes}
\usepackage{tabularx}
\usepackage{gensymb}
\usepackage[utf8]{inputenc}
\usepackage{mathtools, nccmath}
\usepackage{subcaption}
\usepackage{booktabs}
\usepackage{cleveref}
\usepackage{chemformula}
\usepackage[version=4]{mhchem}
\usepackage{booktabs, multirow} 
\usepackage{soul}
\usepackage{tablefootnote}
\modulolinenumbers[1]

\usepackage{algorithm}
\usepackage{algpseudocode}
\biboptions{authoryear}
\definecolor{DarkGreen}{rgb}{0.2,0.5,0.2}
\hypersetup{
    colorlinks=true, 
    citecolor=DarkGreen,  
    linkcolor=black, 
    filecolor=black,
    urlcolor=blue    
}

\usepackage{setspace}

\usepackage{comment}
\usepackage[inline]{enumitem}   
\makeatletter
\newcommand{\inlineitem}[1][]{%
\ifnum\enit@type=\tw@
    {\descriptionlabel{#1}}
  \hspace{\labelsep}%
\else
  \ifnum\enit@type=\z@
       \refstepcounter{\@listctr}\fi
    \quad\@itemlabel
    \hspace{\labelsep}%
\fi}
\makeatother
\begin{document}
\begin{frontmatter}

\title{Deep reinforcement learning-based longitudinal control strategy for automated vehicles at signalised intersections}



\author[mymainaddress]{Pankaj Kumar}
\author[mymainaddress]{Aditya Mishra}
\author[mymainaddress]{Pranamesh Chakraborty}\corref{mycorrespondingauthor}
\author[mymainaddress1]{Subrahmanya Swamy Peruru}

\cortext[mycorrespondingauthor]{Corresponding author\\
pranames@iitk.ac.in}

\address[mymainaddress]{Department of Civil Engineering, Indian Institute of Technology Kanpur, Kanpur-208016, U.P., India}
\address [mymainaddress1]{Department of Electrical Engineering Indian Institute of Technology Kanpur, Kanpur-208016, U.P., India}

\begin{abstract}
Developing an autonomous vehicle control strategy for signalised intersections (SI) is one of the challenging tasks due to its inherently complex decision-making process. This study proposes a Deep Reinforcement Learning (DRL) based longitudinal vehicle control strategy at SI. A comprehensive reward function has been formulated with a particular focus on (i) distance headway-based efficiency reward, (ii) decision-making criteria during amber light, and (iii) asymmetric acceleration/ deceleration response, along with the traditional safety and comfort criteria. This reward function has been incorporated with two popular DRL algorithms, Deep Deterministic Policy Gradient (DDPG) and Soft-Actor Critic (SAC), which can handle the continuous action space of acceleration/deceleration. The proposed models have been trained on the combination of real-world leader vehicle (LV) trajectories and simulated trajectories generated using the Ornstein-Uhlenbeck (OU) process. The overall performance of the proposed models has been tested using Cumulative Distribution Function (CDF) plots and compared with the real-world trajectory data. The results show that the RL models successfully maintain lower distance headway (i.e., higher efficiency) and jerk compared to human-driven vehicles without compromising safety. Further, to assess the robustness of the proposed models, we evaluated the model performance on diverse safety-critical scenarios, in terms of car-following and traffic signal compliance. Both DDPG and SAC models successfully handled the critical scenarios, while the DDPG model showed smoother action profiles compared to the SAC model. Overall, the results confirm that DRL-based longitudinal vehicle control strategy at SI can help to improve traffic safety, efficiency, and comfort.

\end{abstract}


\end{frontmatter}
\section{Introduction}
\label{sec: introduction}

Autonomous driving technology holds significant promise for enhancing road safety by minimising the crash risk \citep{gao2021autonomous}. This can be primarily attributed to the uniform decision-making criteria in autonomous vehicles (AVs), unlike human behaviour, which has inherent stochasticity. Further, improved sensor technologies used in AVs for accurate sensing of the surrounding environment can reduce human perception errors. Traditionally, AV design has been based on developing decision-making criteria using sensor data information, where the backend models are primarily trained on human-driven vehicle data. However, studies have shown that human driving can be suboptimal \citep{hart2024towards}. For example, while manoeuvring through a traffic intersection, human drivers primarily pay attention to safety (i.e., avoiding collisions) and urgency to reach the destination. While balancing these two criteria, human drivers often fail to consider other potential improvement regions, such as fuel efficiency. Therefore, training AV models using \textit{only} human driving data limits their efficiency to the human performance level. 

Reinforcement learning (RL) provides an alternate approach to model AV movement using a reward function-based strategy. In this approach, the model is trained with the objective of maximising rewards, instead of active supervision by human driving data. Therefore, it is evident that designing the reward function is one of the fundamental tasks while creating an RL-based AV model. Several past studies have focused on developing Deep Reinforcement Learning (DRL) based vehicle movement models. This includes studies related to typical car-following (CF) behaviour in highways, lane-changing models, merging in on-ramps and diverging in off-ramps, unsignalised intersection manoeuvring, and roundabouts. A detailed, comprehensive review of DRL-based studies for vehicle manoeuvring is available in \cite{he2017integrated, zhou2019development, zhu2020safe, guo2021hybrid, okhrin2022simulating, yang2024eco, hart2024towards}. Although a significant number of studies have focused on various aspects of developing robust DRL models for car-following in highway scenarios, there has been limited attention on the development of DRL models for manoeuvring through signalised traffic intersections. This study attempts to fill the existing gaps in DRL-based research on longitudinal movement of an AV at signalised intersections (SI).

A typical through movement at a signalised traffic intersection consists of two major components: (a) traffic signal compliance and (b) safely following a leader vehicle (LV), if present. The objective here is to control the longitudinal movement of the ego vehicle (also known as an agent in terms of RL and as a Follower Vehicle (FV) in terms of CF behaviour). For through movement at a signalised traffic intersection, the agent's motion can be safely assumed to be controlled by its acceleration or deceleration, which is also known as the actions of the agent. When an LV is present, the agent's behaviour is similar to typical CF behaviour. Several past studies have focused on developing DRL-based CF behaviour. Here, one of the primary objectives of the ego vehicle is to balance between safety (avoiding collision with the LV) and efficiency (increasing the number of vehicles flowing freely on a unit length of the road). Typically, DRL-based CF models have incorporated the efficiency aspect by designing a reward function that encourages the ego vehicle to maintain a specific time-headway gap, obtained from real-world data \citep{zhu2020safe}. However, the time headway value increases rapidly as the vehicle speed approaches zero, a common phenomenon at SI.  This can be handled by designing the reward based on ``\textit{desired distance headway}", which has been used in traditional CF models \citep{treiber2000congested}. Here, the desired distance headway is defined as a function of ego-vehicle speed, thereby allowing it to handle stop-and-go traffic too.

In addition to safely following the LV, the other objective of the ego-vehicle at SI is to comply with the traffic signal rules. While existing DRL-based studies at SI have extensively addressed this by designing a reward function for red and green lights \citep{li2020analysis, wegener2021automated,yang2024eco}, very few studies have extended it to incorporate amber light conditions too \citep{guo2021hybrid}. However, it is imperative that developing decision-making criteria during the amber light is of utmost importance due to the inherent dilemma of whether to stop or go during the amber light. Therefore, it is critical that DRL-based studies handle ego-vehicle movement for amber light conditions, along with regular red and green lights.

Finally, existing DRL-based CF studies have not incorporated the asymmetric response behaviour while developing the reward function. In traditional CF literature, it has been well documented that the response of the ego vehicle to the stimuli is asymmetric \citep{treiber2000congested}. This means that the response of drivers (in terms of the absolute value of acceleration and deceleration) is different for an increase in distance headway (DH) compared to a decrease in DH. This is primarily because the decrease in DH can lead to safety hazards, thereby requiring immediate attention of FV, unlike an increase in  DH. Since human beings are the ultimate occupants of AVs, it is important that AVs are designed in a way that incorporates these kinds of typical CF behaviour.

Therefore, this study proposes a DRL-based vehicle movement model development with the following major contributions: (1) Developing a distance headway-based reward that can handle the stop-and-go behaviour of traffic movement at SI. (2) Developing a reward function for traffic light compliance, particularly focusing on decision-making criteria during the amber light. (3) Incorporating the asymmetric response of the ego-vehicle to stimuli, a critical aspect of human CF behaviour, through an appropriate reward function. (4) Evaluating the proposed DRL model on both real-world and simulated safety-critical scenarios, which can help to demonstrate the robustness and efficiency of the proposed model.   

The structure of this paper is organised as follows. In Section~\ref{sec: literature}, we review the existing studies and recent advancements related to deep learning and RL algorithms. Section~\ref{sec: methodology} outlines the proposed framework, detailing the formulation of the state space, action set, and reward function used in our approach. Section~\ref{sec: algorithm} describes the specific DRL algorithm implemented in this paper. Section~\ref{sec: data and setup} provides the data used in this study, including the real-world and simulated trajectory datasets utilised for training and evaluation. Section~\ref{sec: results_discussion} includes training results and a comprehensive analysis of the performance of the proposed method across different driving scenarios, using both real-world and simulated data. Finally, Section~\ref{sec: conclusion} summarises the findings and contributions of the paper and discusses potential directions for future research.

\section{Related Works} 
\label{sec: literature}
This section provides a comprehensive review of existing literature on developing models for longitudinal driving of AV from the perspective of supervised deep learning and DRL-based studies. In recent years, the field of deep learning has grown tremendously, relying on supervised Deep Neural Networks (DNN) to process complex driving tasks from input data collected from human driving. Several researchers have applied such supervised imitation learning techniques for modelling longitudinal driving. For instance, \cite{chong2011simulation, khodayari2012modified} used DNNs to predict the FV acceleration for given input as relative speed, intended speed, and gap. Additionally, recurrent neural networks (RNNs) have been utilised to capture temporal dependencies in longitudinal driving \citep{zhou2017recurrent, wang2017capturing}. These supervised machine learning approaches have been shown to achieve smooth, adaptive acceleration and deceleration in longitudinal driving, however, this approach has certain limitations. These techniques typically require a large amount of high-quality naturalistic or simulated data to train the
 models effectively. Poor generalisation capabilities and network susceptibility to bias in decision-making, particularly when the datasets are small and limited driving scenarios that were not present in the training data. Additionally, these models primarily aim to emulate the human driver’s output (acceleration/deceleration and spacing) by adjusting parameters to optimise acceleration and spacing, ensuring that the predicted actions align closely with those in the calibration dataset.
 
To overcome these issues, deep reinforcement learning (DRL) has gained significant attention in the research community in recent years for modelling the ego vehicle action as an RL agent. 
Several studies have explored the potential of DRL in various aspects of automated driving. DRL models have been used to develop and train car-following models for highways and urban roads, including signalised and unsignalised intersections. 
For highways, existing studies aim to learn longitudinal control of ego vehicles by optimising acceleration and deceleration strategies in response to LVs. They formulate the reward function by incorporating key driving parameters, which include safety, efficiency, and comfort \citep{zhou2019development, zhu2020safe, bai2022hybrid, okhrin2022simulating, hart2024towards}. Specifically, Time-To-Collision (TTC) has been used as a safety metric, while time headway and jerk have been considered as indicators of driving efficiency and comfort, respectively. \cite{zhou2019development} used a log-normal-based probability distribution curve to formulate the time headway-based efficiency reward function, where the distribution parameters were extracted from the study data. However, this approach has two limitations. First, although time headway-based rewards can work for highway conditions, they cannot be extrapolated to stop-and-go conditions, typically observed in urban roads. This is primarily because the time headway value increases rapidly when the vehicle speed approaches zero, thereby creating difficulties in comparing the time headway obtained from the DRL model with the desired time headway obtained based on study data. Further, the desired time headway is kept constant for all speed conditions and calculated directly from the study data, limiting its adaptability for diverse highway speed conditions. Since car-following is an integral part to consider for developing an RL agent in the SI, it is of utmost importance to develop the efficiency reward function to handle the stop-and-go traffic of SI too.

Along with car-following behaviour, modelling a DRL agent's longitudinal movement in SI requires developing the reward function for traffic signal compliance. Limited studies have focused on developing DRL models for SIs. Most of such studies have focused on developing eco-driving strategies (i.e., improving fuel efficiency) to manoeuvre the traffic intersection. \cite{wegener2021automated} and \cite{li2022deep} developed DRL models for SI using the Twin Delayed Deep Deterministic Policy Gradient (TD3) model. Similarly, \cite{bai2022hybrid} proposed a hybrid RL framework for learning the movement of connected and autonomous vehicles in SI. However, these TD3-based models used sparse reward functions for violating traffic signals and collision only, ignoring headway-based efficiency rewards. Further, all these studies used acceleration values obtained from car-following models (e.g. Intelligent Driver Model or Krauss car following model) directly as the desired action values, thereby limiting the RL model to learn the optimum action based on a trial-and-error strategy. On the other hand, \cite{yang2024eco} proposed a DRL model for vehicle manoeuvre in SI, specifically developing a reward function for traffic signal compliance. However, their reward function considered green and amber signals equivalently, thereby ignoring the critical decision-making process involved with vehicle movement during the amber light. On the other hand, \cite{guo2021hybrid} handled vehicle movement during amber light by comparing the ego vehicle speed with the desired speed obtained from the Krauss CF model \citep{guo2021hybrid}. Overall, there is a need to develop RL models taking into consideration both efficiency and traffic signal compliance, with particular focus on the decision-making process during the amber light.  

Along with these limitations in DRL models for SI, as discussed above, there are also some other critical issues that have not been addressed in the literature. The majority of the existing studies have assumed that acceleration and deceleration ranges are symmetric. The commonly used ranges for deceleration and acceleration are $-3\,\text{m/s}^2$ to $3\,\text{m/s}^2$ \citep{zhou2017recurrent, zhu2020safe, bai2022hybrid} and $-4\,\text{m/s}^2$ to $ 4\,\text{m/s}^2$ \citep{wu2019deep, yang2024eco}. 
We found only one study that explicitly considered asymmetric acceleration range $-5\,m/s^2$ to $2\, m/s^2$ \citep{hart2024towards}. However, in real-world driving, acceleration and deceleration ranges are expected to be asymmetric in nature. For instance, the comfortable acceleration rate falls within the range of $0\, \text{m/s}^2$ to $2\,\text{m/s}^2$, while comfortable deceleration is around $-4\,\text{m/s}^2$ \citep{treiber2000congested}. In a real driving scenario, drivers decelerate more aggressively to avoid collisions than they accelerate when the LV moves farther away. This imbalance is crucial for modelling realistic car-following behaviour, particularly in urban traffic (stop-and-go traffic) conditions. Furthermore, none of the existing studies take into account the asymmetric acceleration and deceleration capabilities by developing a suitable reward function that can help the RL agent to learn this behaviour.  

Finally, while training the DRL models, the majority of the studies have used either real-world datasets such as NGSIM \citep{usdot_ngsim_2009}, HighD \citep{krajewski2018highd}, pNEUMA \citep{barmpounakis2020new} or synthetic driving simulation and driving cycle data \citep{wu2019deep, qu2020jointly, guo2021hybrid, wegener2021automated}. Only a few studies have explored the integration of both real-world and simulated data for training the RL model \citep{li2023modified, hart2024towards} and none of them are focused on SI. The key limitation of the real-world dataset is the lack of safety-critical scenarios where maximum deceleration may be necessary. Therefore, relying solely on real-world data may lead to RL models that can perform poorly in risky situations. Along with training, the existing RL-based studies on SI have often overlooked testing the developed model for different safety-critical situations, such as the sudden harsh braking of LV or ego vehicle movement towards the end of green or amber phases. It is extremely important that the proposed models be tested for diverse safety-critical scenarios to ensure robustness of the models. These situations are critical in real-world driving, particularly for stop-and-go traffic at SI, which can significantly impact traffic safety and collision risk. 

Overall, the proposed DRL model in this study attempts to fill in the existing research gaps by considering efficiency and traffic signal compliance simultaneously, along with asymmetric acceleration/deceleration response behaviour and finally training and testing the models on both real-world and simulated data to ensure the robustness of the model. In the next section, we discuss the details of the reward function of the proposed DRL model. 

\section{Proposed framework}
\label{sec: methodology}
This section presents the proposed RL-based vehicle control framework designed for navigating a signalised traffic intersection. The methodology is structured into three key components. Section~\ref{subsec: state_action} explains the representations of states, actions, and state update details. Section~\ref{subsec: rewrad_formulation} discusses the reward function formulation, which serves as the guiding criterion for optimising the agent's learning process. Finally, Section~\ref{subsec: collision_avoidance} describes the implementation of action masking for collision avoidance. 


\subsection{States and action representation} 
\label{subsec: state_action}
\begin{figure}[htbp]
    \centering
    \includegraphics[width=0.9\linewidth]{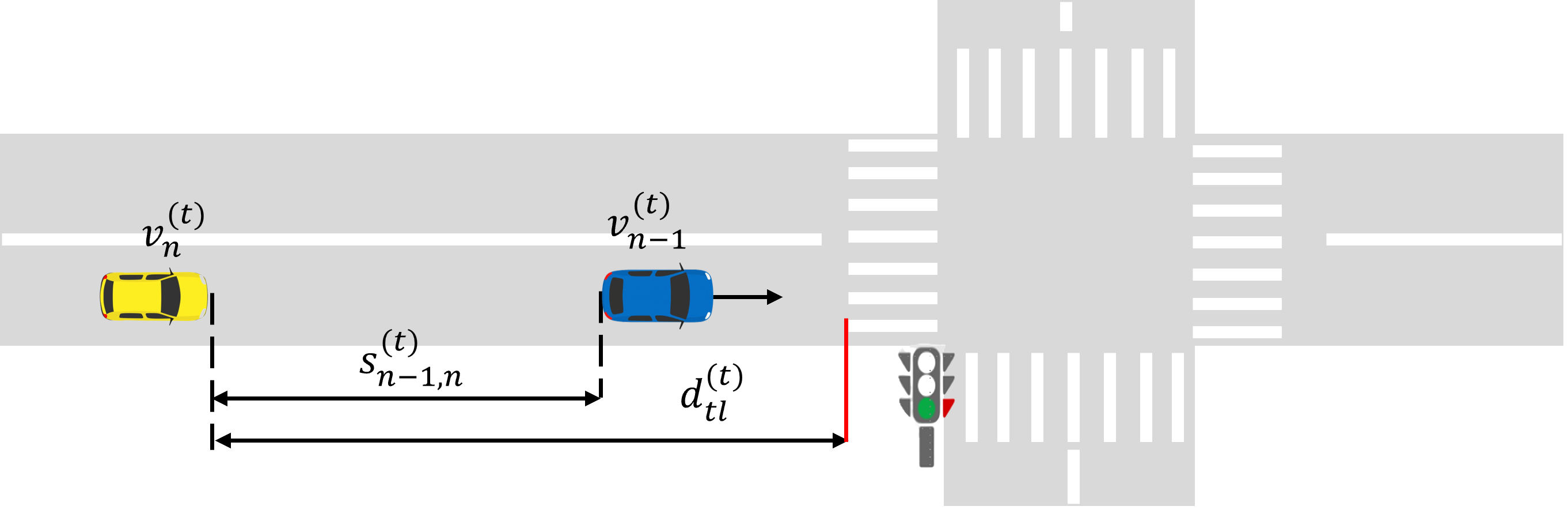}
    \caption{Overview of state representation used in proposed RL framework}
    \label{fig: state}
\end{figure}

The primary objective of this study is to control the agent (ego vehicle) during through movement while manoeuvring at SI, shown as the yellow vehicle in  Fig.~\ref{fig: state}. When another vehicle is present in front of the ego vehicle, the ego vehicle is referred to as the follower vehicle (FV), while the vehicle ahead is termed as the leader vehicle (LV) (shown as blue vehicle in Fig.~\ref{fig: state}).
The ego vehicle chooses action depending on the set of influencing factors, including the behaviour, presence of the LV, and the state of the traffic signal. These influencing factors are referred to as the state space, providing the agent with the necessary observations to make optimal driving decisions. Fig.~\ref{fig: state} shows a pictorial representation of the state variables used in our proposed RL framework. The ego vehicle (FV) and the LV are represented as vehicle \(n\) and \((n-1)\) respectively. The state variables used to represent the influence of the traffic signal are (1) Distance of ego vehicle from the stop line $(d_{tl}^{(t)})$ and (2) Traffic light status (i.e., red/yellow/green) represented as TL. Additionally, when a LV is present, the state variables used to represent the influence of the LV are (1) Spacing between the LV and the FV denoted as $ S_{n-1,n}^{(t)}$, (2) Speed of ego vehicle $v_n^{(t)}$, (3) relative velocity \( \Delta v_{n-1,n}^{(t)} \) which represent the difference between the speed of LV and FV $(v_{n-1}^{(t)}-v_n^{(t)})$.
Overall, Eq.~\eqref{eq: state} represents the state vector at time step $t$. Based on these states at each time step $(t)$, the RL agent selects an action to optimise performance. The action space is continuous and consists of one-dimensional longitudinal acceleration/deceleration of the ego vehicle. This action is defined by the range of possible accelerations \({a}^{(t)} \in [-a_{\text{min}}, a_{\text{max}}] \). where $a_{min}$ and $a_{max}$ represents the maximum allowable deceleration and acceleration limits respectively.


The selection of state and action variables in this study is inspired by previous RL studies, particularly in the context of car-following scenarios \citep{mnih2015human, zhou2019development, zhu2020safe, hart2024towards} and traffic signal control \citep{guo2021hybrid, yang2024eco}.

\begin{equation}
s^{(t)} = \left(v_n ^{(t)}, \, \Delta v_{n-1,n} ^{(t)}, \,  S_{n-1,n} ^{(t)}, \,   d_{tl} ^{(t)}, \, TL^{(t)} \right)
\label{eq: state}
\end{equation} 

\subsubsection*{State update}
The state vector $s^{(t)}$ needs to be updated for each timestep based on the action taken by the agent and the information of LV. In this study, we have considered the agent, i.e. ego vehicle as a point mass. Consequently, its state is updated using point mass kinematics.
At each time step $t$, the vehicle's speed and position are updated for the next time step using Eq.~\eqref{eq: state_update}. This kinematic model has been widely used in vehicle-following \citep{treiber2015comparing} and RL studies \citep{mnih2015human} due to its effectiveness in simulating realistic driving while maintaining computational efficiency.

\begin{equation}
\begin{aligned}
v_n^{(t + 1)} &= v_n ^{(t)} + a_n^{(t)} \times \Delta T \\
\Delta v_{n-1,n}^{(t + 1)} &= v_{n-1}^{(t + 1)} - v_n^{(t + 1)} \\
S_{n-1,n}^{(t + 1)} &= S_{n-1,n}^{(t)} + \Delta v_{n-1,n}^{(t)} \times \Delta T \\
d_{tl}^{(t+1)} &= d_{tl}^{(t)}+v_n^{(t+1)} \times \Delta T + 0.5 \times a_n^{(t+1)} \times (\Delta T)^2\\
TL^{(t+1)} &= \text {Traffic signal phasing}
\end{aligned}
\label{eq: state_update}
\end{equation}

Where, $v_n ^{(t)}$ and $v_n ^{(t+1)}$ denote the ego vehicle's speed at time steps $t$ and $t+1$ respectively, while, $\Delta T$ is the interval between two consecutive simulation time steps, set as 0.04 sec in this study. The leader vehicle's speed $(v_{n-1}^{(t+1)})$ is obtained directly from the data (real-world or simulated trajectory) at each time step. The details of the real world and simulated trajectory data is discussed later in Section~\ref{sec: data and setup}. This ensures that the ego vehicle's response is based on LV trajectories. The initial state of the ego vehicle is directly taken from the real-world or simulated trajectory data to ensure that the ego vehicle starts from a realistic scenario. This prevents arbitrary or unrealistic initial conditions that might not reflect real driving behaviour. The environment simulates subsequent states of the ego vehicle, which is responsible for updating the vehicle's next states using the relations given in Eq.~\eqref{eq: state_update}. The environment takes the current state as input, chooses an appropriate action (acceleration, deceleration), and then updates the state accordingly using the kinematic equations.

\subsection{Rewards function formulation} 
\label{subsec: rewrad_formulation}
The formulation of the reward function is the most important element of the RL-based vehicle control model, as it directly influences the agent's learning processes. It motivates the agent to take appropriate action for the given states while discouraging undesirable actions. In the context of longitudinal vehicle control at SI, the vehicle response can be divided into two primary components: compliance with traffic signals and regular car-following behaviour. Traffic compliance involves obeying the traffic signal and speed limit rules. Along with this, the ego-vehicle also follows typical car-following behaviour when LV is present in front of the ego-vehicle.
On the other hand, car-following behaviour represents how the FV (same as the ego vehicle) adjusts the movement based on the LV behaviour \citep{pipes1953operational, gaizs1961non, treiber2000congested}. Car-following involves three key driving objectives: safety, efficiency, and comfort. In the next section, we discuss the reward function design related to different objectives that the ego-vehicle tries to balance for compliance with traffic signal rules and following the LV.

\subsection*{Compliance with traffic light regulations}  
\label{subsec: traffic_light_rule}
The main objective of this reward function is to encourage the vehicle to make appropriate decisions (stop or go) during the amber phase while strictly enforcing stopping behaviour at a red light. The function must penalise unsafe or abrupt action, such as (a) trying to stop at the traffic light signal, which would be impossible even with maximum deceleration, or (b) attempting to cross when stopping is the safer choice.

\begin{figure}[htbp]
    \centering
    \includegraphics[width=0.9\linewidth]{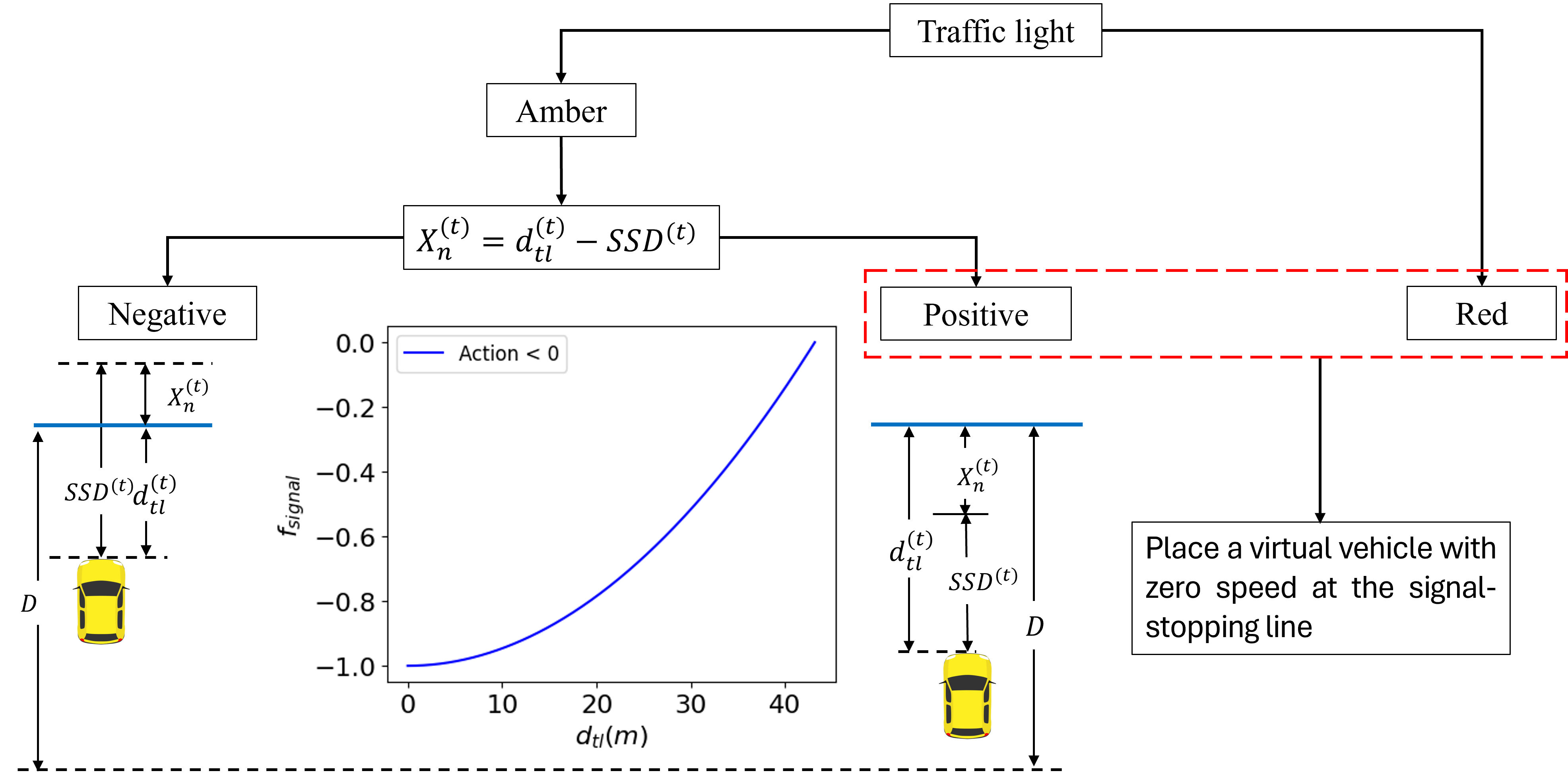}
    \caption{Illustration of vehicle behaviour and penalty conditions when approaching to traffic light}
    \label{fig: signal}
\end{figure}

Fig.~\ref{fig: signal} illustrates the key variables associated with the traffic signal compliance reward function, specifically the amber and red signal phases. Fig.~\ref{fig: signal} is divided into two parts: (a) amber light situation and (b) red light situation. Please note that no separate reward is required for green conditions because the ego vehicle is expected to continue with the same speed during green light conditions unless a leader vehicle is present. If LV is present, then it will lead to car-following-based rewards, which we will discuss in the next reward section.

In Fig.~\ref{fig: signal}, the ego vehicle is represented using the yellow vehicle icon, and the blue line shows the stop line where the vehicle needs to stop during red light conditions. $D$ is the distance between the stop line and the end of the signal influence area (shown as a dotted line), representing the zone where the ego vehicle needs to take action based on the state of the traffic signal. $d_{tl}^{(t)}$ represents the distance of the ego vehicle from the stop line at time $t$, while $SSD^{(t)}$ denotes the stopping sight distance of the vehicle at time $t$. The ego vehicle needs to make a decision whether to stop or go based on the minimum distance required for the vehicle to come to a stop, denoted using $d_{tl}^{(t)}$, as shown in Eq.~\eqref{eq: state_update}.


\begin{equation}
SSD^{(t)} =  v_{n}^ {(t)} T_r + \frac{v_{n}^ {(t)(2)}} {2a_{comf}}
\label{eq: SSD}
\end{equation} 

where $v_n^{(t)}$ is the current speed, $T_r$ represent the reaction time, and $a_{comf}$ is the comfortable deceleration. 
The variable $X_n^{(t)}$ represents the difference between $d_{tl}^{(t)}$ and $SSD^{(t)}$ and determines whether the ego vehicle can safely stop or not during an amber light depending upon its sign, which can be positive or negative. When the traffic light turns to amber, the agent will first decide whether it can cross the intersection or needs to stop based on the value of $X_n^{(t)}$.
When $X_n^{(t)}\leq 0$, the vehicle cannot stop before the stop line even when applying maximum comfortable deceleration. In this case, if the agent decelerates, it will incur a penalty as defined in Eq.~\eqref{eq: f_signal}.
On the other hand, when $X_n^{(t)}>0$, it indicates that $d_{tl}^{(t)}$ is greater than $SSD$, meaning the vehicle has enough room to stop safely. Therefore, in this scenario, we place a standing virtual vehicle of zero length at the stop line, which can be removed when the traffic signal turns green. This allows us to model the deceleration behaviour of an ego vehicle similar to a stalled LV instead of separately creating a reward function specific to stop at a traffic light. In traditional CF literature and RL-based signal rewards \citep{gao2021autonomous}, a similar assumption of simulating a red light by a standing virtual vehicle has been used \citep{treiber2013traffic}.

The amber light decision-making process has been expanded upon in the red light scenario. When the traffic light turns red, a stalled virtual vehicle of zero length is placed at the signal stop line, which will be removed when the signal light turns green. The reward to handle the interaction between the stalled LV and the ego-vehicle (also popularly called the follower vehicle (FV) in car-following literature) will be discussed later during the \textit{efficiency} rewards formulation. The overall reward function to handle the traffic signal compliance for the ego-vehicle is given by Eq.~\eqref{eq: f_signal}. Please note that $f_{Eff}$ is the reward for efficiency which be discussed later. 

\begin{equation}
f_{\text{Signal}}^{(t)} =
\begin{cases} 
    \left( \frac{d_{tl}^{(t)}}{SSD ^{(t)}} \right)^{2}-1 & \text{TL}^{(t)} = amber, \, X \leq 0, \, a < 0, \, LV_{pos} > \text {Stopping line} \\
    f_{Eff}^{(t)} & \text{TL}^{(t)} = amber, \, X > 0, \, \text{Place virtual vehicle of zero speed at stopping line} \\
    f_{Eff}^{(t)} & \text{TL}^{(t)} = red, \, \text{Place virtual vehicle of zero speed at stopping line}
\end{cases}
\label{eq: f_signal}
\end{equation}

The denominator in Eq.~\eqref{eq: f_signal}, $SSD$, has been used to normalise the reward range so that the different rewards can be combined easily using appropriate weights. To encourage greater sensitivity to smaller instantaneous $d_{tl}^{(t)}$ values, which are more desirable, an exponent of ${2}$ is applied.

\subsection*{Ensuring safety}
\label{subsec: ensure safety}
One of the primary requirements of the ego vehicle during through movement at a traffic intersection is to maintain safety, i.e., not collide with the leader vehicle, irrespective of the actions of the LV. The TTC index has been widely used in traffic safety-related studies to evaluate rear-end collision risk (\cite{li2020analysis}). TTC is defined as the time taken by the following ego vehicle to collide with the LV if neither of them changes their current driving state (i.e., speed or lane). It has also been widely used to model safety criteria for following vehicles in past RL-based car-following models \citep{zhou2019development, zhu2020safe, li2020analysis, guo2021hybrid, yang2024eco}. A smaller TTC value indicates higher collision risk, i.e., TTC and collision risk are inversely related. To mitigate this risk, the ego vehicle adjusts the acceleration or deceleration accordingly. Mathematically, TTC is computed using the following relation: 
\begin{equation}
TTC^{(t)} = -\frac{S_{n-1,n}^{(t)}}{\Delta v_{n-1,n}^{(t)}}
\label{eq: TTC}
\end{equation}

Please note, throughout this study, the agent (FV) has been taken as a point mass, and hence, the length of the vehicle is not considered. However, the methodology can be easily extended to include the vehicle length for computation whenever necessary. $S_{n-1, n}^{(t)}$ denotes the distance between LV and FV while, $\Delta v_{n-1, n} ^{(t)}$ represents relative speed shown in Fig.~\ref{fig: state}. When the relative speed is negative (i.e., the FV moves at a lower speed than the LV), then there is no chance of rear-end collision. Therefore, the situation of concern arises when the relative speed or TTC is positive. \cite{vogel2003comparison} has reported that the TTC threshold can vary from $1.5 sec$ to $5 sec$. In this study, $4$sec has been considered as the TTC threshold similar to \cite{yang2024eco}. Accordingly, the ego vehicle is penalised when the TTC is less than the TTC threshold. The corresponding reward function is given by Eq.~\eqref{eq: f_TTC}. This reward function has also been used in previous RL-based CF models for evaluating TTC-based safety criteria \citep{zhou2019development,zhu2020safe, yang2024eco}.

\begin{equation}
f_{\text{TTC}}^{(t)} = \begin{cases} \left( \frac{TTC ^{(t)}}{TTC_{\text{threshold}}} \right)^2 - 1 & \text{if } 0 \leq TTC^{(t)} < TTC_{\text{threshold}} \\0 & \text{otherwise}
\end{cases}
\label{eq: f_TTC}
\end{equation}

As it can be seen from Eq.~\eqref{eq: TTC}, when $TTC^{(t)}$ is less than $TTC_{threshold}$, $f_{TTC}^{(t)}$ will be negative, resulting in the reduction of the overall reward. The lesser the TTC value, the more negative the reward will be. If TTC approaches zero, the $f_{TTC}^{(t)}$ will be the maximum penalty value, -1, therefore heavily penalising the near-crash situations.

Despite using the penalty function to avoid low TTC, the RL model's action can still lead to a collision, particularly at the initial stages of model training. Since the collision is the worst scenario for any vehicle and must be avoided at any cost, therefore, a significantly large penalty should be given whenever a collision occurs. This can help the RL model to understand how critical it is to avoid a collision. Accordingly, a penalty term of $-50$ has been used in the event of a collision. It can be noted that the range of the rewards/penalties for all other criteria used in this study ranges between $+1$ and $ -1$. Accordingly, the value of $-50$ is chosen for collision. To justify the importance of collision avoidance, the reward is shown in Eq.~\eqref{eq: f_col}. Previous RL models have also chosen similar large penalty values for collision scenarios. The value $50$ was chosen in this study after iterating with different weight values for different rewards. Additionally, the episode is terminated when a collision occurs.

\begin{equation}
f_{\text{Col}}^{(t)} = 
\begin{cases} -50 & \text{if collision occurs} \\ 0 & \text{otherwise}
\end{cases}
\label{eq: f_col}
\end{equation}

\subsection*{Distance headway}
The TTC reward penalises low TTC values, encouraging the vehicle to keep a larger headway.
This can lead to a situation where the FV (i.e., the agent) maintains a very large headway with the LV, which reduces the number of vehicles in the unit length of the road, therefore impacting the efficient usage of road space. Therefore, a balance needs to be maintained between safety and efficiency conditions.

Traffic movement efficiency can be increased by reducing the gap between individual vehicles, thereby incorporating more vehicle movement per unit of time or distance. The gap between the LV and the FV can be defined using two different terms: time headway and distance headway (DH) (also referred to as space headway). Time headway is defined as the time difference between the LV and the FV when they cross a specific point, while DH is defined as the distance between the LV and the FV at any given point. These two terms are the basic microscopic variables of traffic flow theory. Previous RL studies have used time headway-based reward function for handling traffic movement efficiency both on highways \citep{zhou2019development, zhu2020safe, okhrin2022simulating, hart2024towards}, and in urban conditions \citep{guo2021hybrid, yang2024eco}. However, as highlighted in Section ~\ref{sec: literature}, time headway is not suitable in urban scenarios where stop-and-go movement is observed. Therefore, in this study, we have proposed a distance headway-based reward for efficiency. 

\begin{figure}[h]
    \centering
    \includegraphics[width=\linewidth]{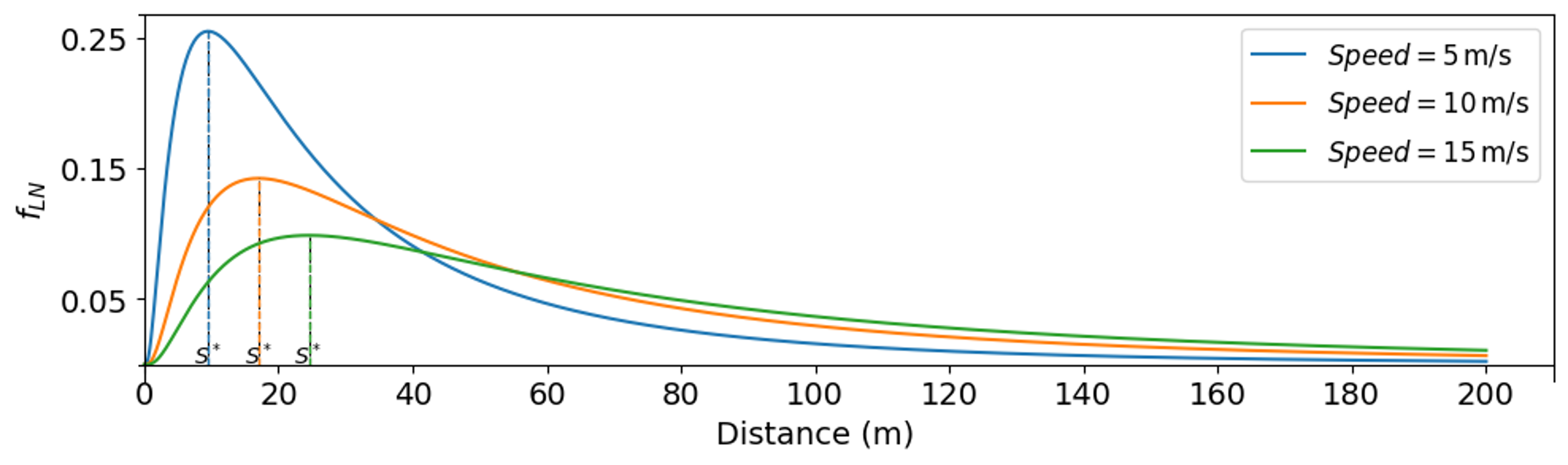}
    \caption{Probability density function (pdf) of a log-normal distribution}
    \label{fig: lognormal}
\end{figure}

Although autonomous vehicles can be modelled to maintain a DH that is ``just" adequate enough to avoid collision (thereby increasing traffic efficiency). However, it can lead to situations where the headway is too close for the human passengers to feel comfortable. It has been well documented in traditional CF literature that human drivers prefer to maintain a ``safe" DH, which is directly proportional to their speed \citep{treiber2000congested, pipes1953operational}.
While modelling RL-based autonomous vehicles, it is important to maintain this safe DH since humans will be the final users of the AV.  Treiber et al. \citep{treiber2000congested} incorporated this by introducing the concept of desired minimum gap $S^*$ defined as shown in Eq.~\eqref{eq: space_headway}.

\begin{equation}
    S^{*(t)}= S_0+v_n^{(t)}T
    \label{eq: space_headway}
\end{equation}
where, $S_0$ is the bumper-to-bumper distance in standing traffic, $T$ is the safe time headway, and $v_n^{(t)}$ denotes the vehicle's current velocity. Eq.~\eqref{eq: space_headway} implies that the desired minimum gap is directly proportional to the follower vehicle's speed. 
Using the concept of the desired minimum gap $S^*$, we have formulated our reward function as the probability density function (pdf) of a log-normal distribution as shown in Fig.~\ref{fig: lognormal}, with the mode, i.e., the peak of the distribution occurring at $S^*$. This is similar to the reward function used by Zhu et al. \citep{zhu2020safe}, where pdf with time headway was used, and the parameters of the distribution were determined from the study data itself. In our study, we used the parameters required for defining the peak $S^*$ (i.e., $s_0$ and $T$) from the study \citep{treiber2000congested} by using Eq.~\eqref{eq: space_headway}. We did not use the calibration parameters from our study data since the real-world dataset used in this study, which will be explained in Section ~\ref{sec: data and setup}, is not a significantly large dataset to empirically determine the best parameter values. In future, a larger real-world dataset can be used to determine the best parameters empirically. The pdf of the log-normal distribution used for the reward function is shown in Eq.~\eqref{eq: lognorm}. It can be noted that selecting a log-normal distribution, which is a right-skewed distribution, as a reward function helps in producing a behaviour where a steeper decrement of reward values occurs at the left of the peak (lower gap than $S^*$) compared to the right side (higher gap than $S^*$). Studies have shown that humans are more hesitant to keep a lower gap with the LV than their "safe gap" since this leads to safety-critical issues \citep{treiber2000congested}. Also, $ S^*$ is defined as the desired \textbf{minimum} gap and hence, conditions where $S < S^*$ should ideally be avoided. The right-skewness of the log-normal reward function helps us to make our RL model understand that keeping a lower gap than $S^*$ is worse compared to keeping a higher gap than $S^*$.

\begin{figure}[h]
    \centering
    \includegraphics[width=\linewidth]{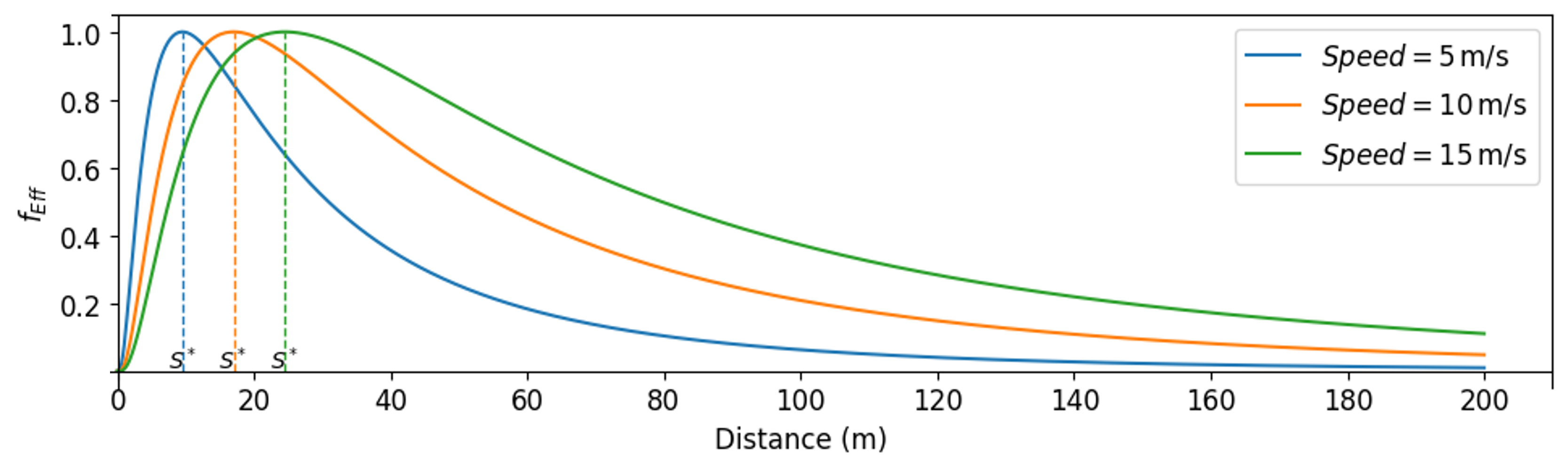}
    \caption{ Reward function for distance headway using a lognormal distribution}
    \label{fig: headway}
\end{figure}

However, the reward function $f_{Eff}^{(t)}$ suffers from one drawback. The peak value of the pdf is dependent on $S^*$, which in turn is dependent on the current speed of FV $(v_n^{(t)})$. Therefore, the maximum reward that the FV can obtain even while maintaining the desired distance headway varies for different $v_n^{(t)}$ as shown in Fig.~\ref{fig: headway}. To handle this issue and to keep the range of reward values of efficiency similar to that of safety, traffic signal, etc., we have normalized the $f_{Eff}^{(t)}$, as shown in Eq.~\eqref{eq: f_eff}. This allows the RL model to get maximum efficiency reward, $+1$, for keeping the distance headway equal to $S^*$, irrespective of the value of $v_n^{(t)}$.


The probability density function of the log-normal distribution is as follows:
\begin{equation}
f_{\text{LN}(S^{(t)}|\mu,\sigma)}^{(t)} = \frac{1}{S^{(t)} \sigma \sqrt{2\mu}} \exp\left(-\frac{(\ln(S^{(t)}) - \mu^2}{2\sigma^2}\right); S^{(t)}>0
\label{eq: lognorm}
\end{equation}
with $\mu = ln(S^{*(t)}) + \sigma^2$, $\sigma = 1$. 
Our final DH reward function is given as follows:
\begin{equation}
    f_{Eff}^{(t)} = \frac{f_{\text{LN}(S^{(t)}|\mu,\sigma)}}{f_{\text{LN}(S^{*(t)}|\mu,\sigma)}}; f_{\text{LN}(S^{*(t)}|\mu,\sigma)}>0
    \label{eq: f_eff}
\end{equation}
where $\mu$ and $\sigma$ denote the mean and standard deviation of the log-normal distribution. $S$ is the DH. $f_{\text{LN}(S^{*(t)}|\mu,\sigma)}$ is employed to normalise the absolute value of each reward component to the interval [0,1]. The $f_{\text{LN}(S^{*(t)}|\mu,\sigma)}$ is calculated by substituting $S= S^*$ in Eq.~\eqref{eq: lognorm}.

\subsection*{Driving comfort}
\label{subsec: driving comfort}
\begin{figure}[h]
    \centering
    \includegraphics[width = 0.9\linewidth]{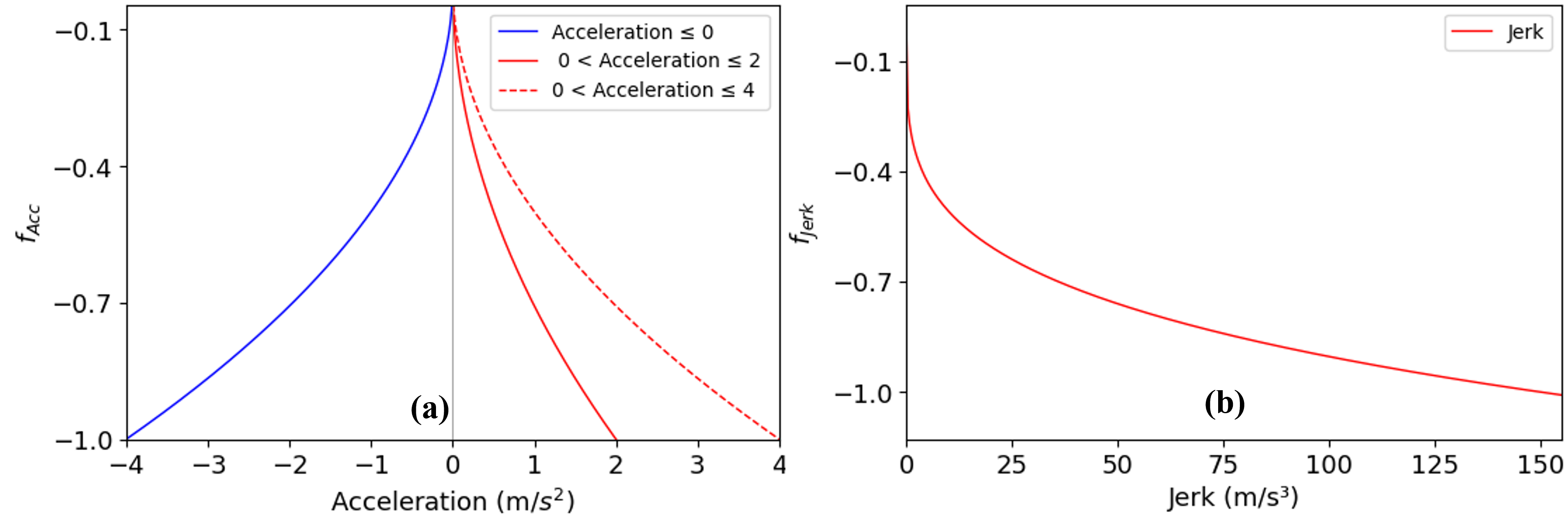}
    \caption{Visualisation of reward function for (a) action and (b) jerk in comfortable driving}
    \label{fig: comfort}
\end{figure}

Ensuring passenger comfort while the vehicle is in motion is a key consideration, particularly in scenarios involving stop-and-go traffic typically observed at SI. Jerk and acceleration are widely recognised indicators of ride comfort and are commonly used to assess how smoothly a vehicle operates. 
\cite{De2023standards} has identified a strong linear correlation between jerk and perceived ride quality. 
This is likely because higher jerk values tend to occur over shorter durations, meaning that abrupt changes in motion thereby minimise discomfort for passengers. This study also showed that higher fluctuations of acceleration and higher values of jerk are the factors that should be reduced to make driving smoother and more pleasant. Based on these, both acceleration and jerk have been added to the formulation of the reward function for comfort driving. 
Similar to the study used by \cite{yang2024eco}, we have used a reward function that considers both jerk and acceleration. However, existing RL-based CF studies have used the maximum absolute value of deceleration, the same as that of maximum acceleration. For example, the range of acceleration values used in \cite{zhou2019development, yang2024eco} was [-4,4], whereas in \cite{zhu2020safe} the range was [-3, +3], respectively. On the other hand, one important property of CF behaviour is the asymmetric response to the stimuli \citep{treiber2000congested}. This means that the driver's response (in terms of absolute value of acceleration and deceleration) to DH decrements is larger compared to increments in DH. This is primarily because the decrease in DH leads to safety hazards where drivers need to immediately respond to such changes, unlike an increase in DH. To incorporate this typical CF behaviour, we used the range of acceleration to be [-4, +2], and the corresponding reward function (in terms of penalty) for large acceleration values is given by  Eq.~\eqref{eq: f_action} and shown in Fig.~\ref{fig: comfort}(a). It can be easily seen that a larger penalty is given for accelerating the ego vehicle compared to deceleration, thereby encouraging the vehicle to have smoother acceleration profiles and avoiding unnecessary large acceleration (and deceleration values).

Along with $f_{Acc}^{(t)}$, the $f_{Jerk}^{(t)}$ handles the comfort of the passenger by imposing a penalty on high jerk values of a jerk, shown in Fig.~\ref{fig: comfort}(b). This equation is similar to the jerk reward used in other RL-based CF studies \citep{yang2024eco}. As it can be seen from Eq.~\eqref{eq: f_action} and Eq.~\eqref{eq: f_jerk}, both $f_{Acc}^{(t)}$ and $f_{Jerk}^{(t)}$ lies between [-1, 0].

\begin{equation}
f_{\text{Acc}}^{(t)} = -\begin{cases} \left( \frac{action}{a_{\text{min}}} \right)^{1/2} & \text{if action }  \leq 0 \\
\left( \frac{action}{a_{\text{max}}} \right)^{1/2} & \text{otherwise}
\end{cases}
\label{eq: f_action}
\end{equation}

\begin{equation}
f_{\text{Jerk}}^{(t)} = -\left[\frac{|j^{(t)}|}{\frac{a_{\text{max}} - a_{\text{min}}}{\Delta T}}\right]^{\frac{1}{4}}
\label{eq: f_jerk}
\end{equation}
\begin{equation}
    f_{Comfort} = f_{Acc}^{(t)} + f_{Jerk}^{(t)}
    \label{eq: f_comfort}
\end{equation}

Here $j$, $a_{min}$, $a_{max}$ and $\Delta T$ denote the jerk, specified maximum deceleration limit $(-4\,m/s^2)$, acceleration limit $(2\, m/s^2)$ and time step respectively. In  Eq.~\eqref{eq: f_comfort},  $\frac{a_{\text{max}} - a_{\text{min}}}{\Delta T}$ is used to normalise the absolute value of each reward component. 

\subsection*{Maintaining desired speed}  
\label{subsec: desired_speed}
Along with compliance with traffic rules, the ego vehicle (i.e., agent) also needs to maintain the speed limit. Therefore, a penalty is imposed when the ego vehicle crosses the predefined speed limit. The reward function for maintaining the desired speed is given by Eq.~\eqref{eq: f_speed}. As shown in Eq.~\eqref{eq: f_speed}, the penalty increases as the difference between the vehicle's speed and the speed limit becomes larger. This discourages excessive speeding by imposing a stricter penalty for higher deviations, ensuring that the ego vehicle adheres to speed regulations and maintains safe driving behaviour. The denominator in Eq.~\eqref{eq: f_speed} is used to normalise the absolute value of the reward term.


\begin{equation}
f_{\text{Speed}}^{(t)} = \begin{cases} - \left(\frac{v_n ^{(t)} - v_{\text{limit}}}{v_{\text{limit}}}\right)^2 & \text{if } v_n^{(t)} > v_{\text{limit}} \\
0 & \text{otherwise}
\end{cases}
\label{eq: f_speed}
\end{equation}

The overall reward function at every simulation time step is formulated as the weighted aggregate of all defined reward components. These components include safety, efficiency, comfort, and compliance with traffic signals and speed limits. The mathematical form of the reward function is expressed as follows:
\begin{equation}
  R^{(t)} = 
  \begin{cases} 
    w_1 f_{\text{Signal}}^{(t)} + w_2 f_{\text{TTC}}^{(t)} + w_3 f_{\text{Eff}}^{(t)} + w_4 f_{\text{Acc}}^{(t)} + w_5 f_{\text{Jerk}}^{(t)}  + w_6 f_{\text{Speed}}^{(t)} & \text{if } S > 0 \\
    w_7 f_{\text{Col}}^{(t)} & \text{otherwise}.
  \end{cases}
  \label{eq: reward}
\end{equation}

Where $w_1$, $w_2$, $w_3$, $w_4$, $w_5$, $w_6$, and $w_7$ are the coefficients of each reward function that determine the relative importance of each reward component. Table~\ref{table: reward_table} lists the chosen value of these coefficients.

\subsection{Collision avoidance strategy} 
\label{subsec: collision_avoidance}
The TTC reward is designed to penalise scenarios where TTC values are low, indicating potential danger. Even after converging the training process with TTC-based rewards, the agent may still take actions that result in unsafe conditions or collisions. In the context of autonomous vehicles, it is extremely important to avoid such critical or hazardous situations. 
To address this issue, this study includes the utilisation of a kinematic-based collision avoidance strategy similar to \citep{zhou2019development, zhu2020safe, guo2021hybrid, yang2024eco}. This approach comprises calculating a minimum safe distance $(d_{safe})$ using Eq.~\eqref{eq: safe_dis}, which is the sum of the distance travelled by the FV (i.e., agent) during the driver's reaction time $T_r$ and braking distance required for the FV to stop under maximum deceleration. Then we check whether the agent action will lead to a distance less than $d_{safe}^{(t)}$. If the projected distance falls below the required safe distance $(d_{safe}^{(t)})$, a predefined safe deceleration $a_{min}$ is applied irrespective of the action given by the agent, to prevent potential collision.
The safe distance $d_{safe}^{(t)}$ is the minimum spacing needed to avoid a collision when the leading vehicle suddenly applies emergency braking while the following vehicle decelerates at its maximum capacity. This distance is calculated as the sum of the reaction and braking distance, as formulated in Eq.~\eqref{eq: safe_dis}.
\begin{equation}
d_{\text{safe}}^{(t)} = v_n^{(t)} T_r + \frac{v_n^{2(t)} - v_{n-1}^{2(t)}}{2 a_{\text{min}}}
\label{eq: safe_dis}
\end{equation}
\begin{equation}
a^{(t)} =\begin{cases} 
    a_{min} & \text{if } S_{n,n-1}^{(t)} < d_{\text{safe}}^{(t)} \\
    \text{model's output} & \text{otherwise}.
\end{cases}
\label{eq: coll_action}
\end{equation}

Here, $T_r$ is the reaction time of FV, and $a_{min}$, $a_{max}$ denote maximum deceleration and acceleration. The chosen value of $T_r$, $a_{min}$, and $a_{max}$ are provided in Table~\ref{table: reward_table}. Next, we discuss the details of the RL framework used in this study to train the model based on the reward functions formulated and described above.

\begin{table}[htb]
    \caption{Hyperparameters utilised for configuring the input states and defining the reward function.}
    \label{table: reward_table}
    \centering
    \renewcommand{\arraystretch}{1} 
    \begin{tabular*}{0.9\textwidth}{@{\extracolsep{\fill}} l l l }
        \toprule
        Symbol & Description & Value \\
        \midrule
        $a_{max}$ & Maximum acceleration &  $2 \, \text{m/s}^2$ \\
        $a_{min}$ & Maximum deceleration  &  $ -4 \, \text{m/s}^2$ \\
        $a_{comf}$ & Comfortable deceleration  & $ -2 \, \text{m/s}^2$ \\
        $v_{des}$ &  Desired speed &  $15 \, \text{m/s}$  \\
        $v_{limit}$  &  Speed limit  & $15 \, \text{m/s}$ \\
        $S_0$   & Bumper-to-bumper distance in standing traffic  & $2 \, \text{m}$ \\
        $T$   & Safe time headway &  $1.5 \, \text{s}$ \\
        $T_r$  & Reaction time of ego vehicle  & $1.5  \, \text{s}$ \\
        $\Delta t$   & Time interval between two consecutive simulation  & $0.04 \, \text{s}$ \\
        $w_1$ & Weight factor for maintaining the traffic signal &  $1$ \\
        $w_2$ & Weight factor for maintaining the safety & $1.5 $ \\
        $w_3$ & Weight factor for maintaining the desired gap  & $1.5 $ \\
        $w_4$ & Weight factor for maintaining the acceleration & $0.3$  \\
        $w_5$ & Weight factor for maintaining the jerk  & $2$  \\
        $w_6$ & Weight factor for maintaining the desired speed &  $2$  \\
        $w_7$ & Weight factor for maintaining the collision &  $50 $ \\        
        \bottomrule
    \end{tabular*}
\end{table}
\section{Reinforcement learning algorithm}
\label{sec: algorithm}
RL is a framework for decision-making where the agent interacts with the environment to learn policy through trial and error to maximise cumulative rewards over time \citep{bellman1957markovian}. The agent follows a policy (maps states to actions) that determines action based on a given state and continuously refines it iteratively using value-based, policy-based, or hybrid approaches. Popular value-based RL algorithms include Q-learning \citep{watkins1992q}, Deep Q-Network (DQN) \citep{mnih2013playing}, State-Action-Reward-State-Action (SARSA) \citep{rummery1994line}, and temporal difference (TD) \citep{sutton2018reinforcement}, which estimate the value function to derive the optimal policy.
These methods work well for situations with distinct action spaces since they leverage prior experiences to improve decision-making. However, the downside of this approach is that it struggles with high-dimensional and continuous action spaces due to value function approximation. 
On the other hand, policy-based methods such as REINFORCE \citep{williams1992simple} and Proximal Policy Optimisation (PPO) \citep{schulman2017proximal} directly optimise the policy by updating parameters based on performance.
These methods are well-suited for continuous action spaces but can be sensitive to hyperparameter tuning and may exhibit high variance in policy updates. To address the limitations of both value-based and policy-based methods, researchers have combined these approaches. Among the most commonly used RL algorithms for continuous control tasks are Deep Deterministic Policy Gradient (DDPG) \citep{lillicrap2015continuous} and Soft-Actor-Critic (SAC) \citep{haarnoja2018soft}.
In this paper, we have used DDPG and SAC algorithms to control vehicle movement at SI, the details of which are discussed in the following section. 

\subsection{Deep Deterministic Policy Gradient (DDPG)}
\label{subsec: DDPG}
DDPG \citep{lillicrap2015continuous} is an off-policy, model-free, and Actor-critic algorithm that combines the concepts of DQN \citep{mnih2015human} and Deterministic Policy Gradient (DPG). The DQN is a value-based method that combines Q-learning with neural networks with target and evaluation models for function approximation. The DQN model has shown the potential to deal with high-dimensional observation spaces. It outputs an action for which we obtain the maximum possible Q-value. When dealing with large state spaces, constructing and following the Q-table would not be feasible. This is when neural networks come in handy. As input, the neural network receives the state and action pair $(s^{(t)}, a^{(t)})$, and provides the Q-value $Q(s^{(t)}, a^{(t)})$ as output. The convergence and stability problems frequently arise when applying RL to address complicated problems because of the correlation between successive states. DQN has the potential to provide a solution to these challenges. On the other hand, the DQN algorithm is only capable of producing discrete actions and, therefore, is not appropriate for continuous actions, such as vehicle movement control problems.\\
The DPG builds on Policy Gradient (PG) \citep{williams1992simple} and the Actor–Critic framework \citep{konda1999actor} to improve continuous action space. PG directly optimises the policy by maximising expected cumulative rewards.
The actor-critic approach combines policy-based (actor) and value-based (critic) learning, where the actor selects actions, and the critic evaluates them using the value function. The actor then modifies its policy based on the critic's feedback, improving action selection over time.  
Both DPG and actor-critic consist of two key components: a critic that updates the value function, while the actor adjusts the policy accordingly.  
Our current study adopted the DDPG learning algorithm with the proposed reward function, discussed in Section~\ref{subsec: rewrad_formulation}. The details of the hyperparameters of the DDPG model used in this study are mentioned in Table~\ref{table: ddpg_table}.

\begin{algorithm}
\caption{Deep Deterministic Policy Gradient (DDPG) Algorithm}
\label{algo: ddpg}
\begin{algorithmic}
\State \textbf{Initialize} critic network \( Q_{\theta}(s^{(t)},a^{(t)}) \) and actor-network \( \pi_{\phi}(s^{(t)}) \) with parameters \( \theta \) and \( \phi \)
\State \textbf{Initialize} target networks \( Q_{\bar{\theta}} \) and \( \pi_{\bar{\phi}} \) with weights \( \bar{\theta} \gets \theta \), \( \bar{\phi} \gets \phi \)
\State \textbf{Initialize} empty replay buffer \( R \)

\For{each episode from \(1\) to \(N\)}
    \State Reset exploration noise \( \epsilon \)
    \State Initialize state \((s_0^{(t)})\) = \(( v_n ^{(t)}, \, \Delta v_{n-1,n} ^{(t)}, \,  S_{n-1,n} ^{(t)}, \,   d_{tl} ^{(t)}, \, TL^{(t)})\)
    
    \For{each timestamp \(t = 1\) to \(T\)}
        \State Select action: \( a^{(t)} = \pi_{\phi}(a^{(t)}|s^{(t)}) + \epsilon^{(t)} \)
        \State Execute action \( a^{(t)} \), observed reward \( r^{(t)} \), and next state \( s^{(t+1)}\)
        \State Store transition tuple \( (s^{(t)}, a^{(t)}, r^{(t)}, s^{(t+1)}) \) in \( R \)
        \State Sample batch from \( R \)
        \State Compute target value: \( y^{(t)} = r^{(t)} + \gamma Q_{\bar{\theta}}(s^{(t+1)}, \pi_{\bar{\phi}}(s^{(t+1)})) \)

        \State Update critic via MSE loss:
        \( L = \frac{1}{N} \sum_{i} \left( y_i^{(t)} - Q_{\theta}(s_i^{(t)}, a_i^{(t)}) \right)^2 \)

        \State Update actor using policy gradient:
        \(\nabla_{\phi} J \approx \frac{1}{N} \sum_{i=1}^{N} \nabla_a Q_{\theta}(s^{(t)}, a^{(t)}) \bigg|_{s=s_i^{(t)}, a=\pi_{\phi}(s_i^{(t)})} \nabla_{\phi} \pi_{\phi}(s^{(t)}) \bigg|_{s=s_i^{(t)}}\)
        \State Soft-update target networks:
        \State \(\bar{\phi} \gets \tau \phi + (1 - \tau) \bar{\phi}\)
        \State \(\bar{\theta} \gets \tau \theta + (1 - \tau) \bar{\theta}\)
        
       
    \EndFor
\EndFor
\end{algorithmic}
\end{algorithm}

\begin{table}[h]
    \caption{Hyperparameters utilised in the DDPG algorithm}
    \label{table: ddpg_table}
    \centering
    \renewcommand{\arraystretch}{1} 
    \begin{tabular*}{0.9\linewidth}{@{\extracolsep{\fill}} l l l }
        \toprule
        Symbol & Description & Value \\
        \midrule
        $LR_A$ & Actor learning rate (ADAM optimiser) & $0.001$ \\
        $LR_C$ & Critic learning rate (ADAM optimiser) & $0.0015$  \\
        $\gamma$ & Discount factor for future reward calculation & $0.99$ \\
        $var_{a}$ & Initial variance of actions for exploration & $3$ \\
        $\lambda$ & Rate of variance reduction for exploration & $0.0005$ \\
        $B$ & Mini-batch size & $256$ \\
        $\tau$ & Soft update rate for target networks & $0.001$ \\
        $R$ & Capacity of the replay memory buffer & $150000$ \\
        $N_l$ & No. of layers in actor and critic networks & $3$ \\
        $N_{hl}$ & No. of neurons in the hidden layers of actor and critic networks & $30$ \\ 
        \bottomrule
    \end{tabular*}
\end{table}

\subsection{Soft-Actor-Critic (SAC)}
\label{subsec: SAC}
SAC is an off-policy RL algorithm that combines policy optimisation with entropy regularisation to enhance exploration \citep{haarnoja2018soft}. It optimised both expected long-term rewards and policy entropy, promoting a balance between exploration and exploitation. The optimal policy $\pi^*$ is derived by resolving:
\begin{equation}
\begin{aligned}
    \pi^* &= \arg\max_{\pi} \sum_{t} \mathbb{E}_{(s^{(t)}, a^{(t)}) \sim \rho_{\pi}} \left[ r^{(t)} + \alpha H(\pi(\cdot | s^{(t)})) \right]
    \end{aligned}
\label{eq: sac_eq1}
\end{equation}

were $\rho_\pi$ denotes the state-action distribution, $r^{(t)}$ is the reward, and $H$ represents the entropy. The entropy term computes as $H(\pi(\cdot|s^{(t)}))$ = $log(\pi_\phi(a^{(t)}|s^{(t)}))$. The temperature parameter $\alpha$ adjusts the entropy weight.  A higher $\alpha$ encourages broader exploration by increasing entropy, leading to more diverse actions, while a lower $\alpha$ results in more deterministic behaviour, balancing exploration and exploitation.
To improve stability and reduce overestimation, the method employs two soft Q-functions, denoted as $Q_\theta$, for evaluating the policy. Each parameter $\theta$ is updated separately by minimising the soft Bellman residual using stochastic gradient descent, as expressed in Eq.~\eqref{eq: sac_eq2}.

\begin{equation}
\begin{aligned}
        \hat{Q}(s^{(t)}, a^{(t)}) &= r^{(t)} + \gamma \mathbb{E}_{s^{(t+1)} \sim \rho_{\pi_s}} \left[ V_{\bar{\theta}}(s^{(t+1)}) \right] \\
        V(s^{(t)}) &= \mathbb{E}_{a^{(t)} \sim \pi} \left[ Q(s^{(t)}, a^{(t)}) - \alpha \log \pi(a^{(t)} | s^{(t)}) \right] \\
        J_Q(\theta) &= \mathbb{E}_{(s^{(t)}, a^{(t)}) \sim R} \left[ \frac{1}{2} (Q_{\theta}(s^{(t)}, a^{(t)}) - \hat{Q}(s^{(t)}, a^{(t)}))^2 \right] \\
    \end{aligned}
\label{eq: sac_eq2}
\end{equation}

were, $R$ denotes the replay buffer, $V_\theta$ is the estimated value function, and $\gamma$ is the discount factor. SAC reduces the risk of overoptimistic value estimates by taking the minimum of the two Q-value estimates $Q_\theta(s^{(t)}, a^{(t)})$, promoting greater stability and reliability in policy learning \citep{haarnoja2018soft}.
The policy improvement is achieved by updating the policy parameters through minimising the expected Kullback-Leibler (KL) divergence while leveraging the reparameterization trick to facilitate efficient optimisation. This enables more efficient gradient computation, and the associated objective function is given in Eq.~\eqref{eq: sac_eq3}.

\begin{equation}
    \begin{aligned}
        J_{\pi}(\phi) &= \mathbb{E}_{s^{(t)} \sim R} \left[ \mathbb{E}_{a^{(t)} \sim \pi_{\phi}} \left[ \alpha \log \pi_{\phi}(a^{(t)} | s^{(t)}) - Q_{\theta}(s^{(t)}, a^{(t)}) \right] \right] \\
    \end{aligned}
    \label{eq: sac_eq3}
\end{equation}

Choosing an appropriate value of $\alpha$ is very difficult as the desired level of entropy may vary throughout learning and between tasks. To address this issue, SAC introduced an automatic temperature-tuning mechanism that helps maintain a balance between exploration and exploitation by minimising a predefined objective function. The temperature parameter optimisation process is described in Eq.~\eqref {eq: sac_eq4}.

\begin{equation}
    \begin{aligned}
        J(\alpha) &= \mathbb{E}_{a^{(t)} \sim \pi_t} \left[ -\alpha \log (\pi_{\phi}(a^{(t)} | s^{(t)})) - \alpha \bar{H} \right]
    \end{aligned}
    \label{eq: sac_eq4}
\end{equation}

The overall SAC training procedure is outlined in algorithm ~\ref{algo: sac}. It starts with the initialisation of the policy network parameters $\phi$, two soft Q-function approximators $Q_1$ and $Q_2$, their corresponding target networks $\bar{Q_1}$ and $\bar{Q_2}$, entropy temperature parameter $\alpha$, and empty replay buffer $R$. Each training episode is initialised with an initial state derived from training data. At each step, an action $(a^{(t)})$ is selected based on the policy $\pi_\phi(a^{(t)}|s^{(t)})$, followed by computing the reward $(r^{(t)})$ and the next state $(s^{(t+1)})$. The resulting transition tuple $(s^{(t)}, a^{(t)}, r^{(t)}, s^{(t+1)})$ is stored in the replay buffer. The algorithm then updates the Q-function, the policy, and temperature parameters using mini-batches sampled from the replay buffer and optimises them via stochastic gradient descent.
The episode terminates when the vehicle’s trajectory ends or a collision occurs. In this study, we also adopted the SAC learning algorithm with the reward function introduced in Section~\ref{subsec: rewrad_formulation}. The specific hyperparameter settings used for the SAC model are provided in Table~\ref{table: sac_table}.

\begin{algorithm}
\caption{Soft Actor-Critic (SAC) Algorithm}
\label{algo: sac}
\begin{algorithmic}
\State \textbf{Initialize} parameters for Q-functions (\(\theta_1, \theta_2\)), policy (\(\phi\)), and temperature (\(\alpha\))
\State \textbf{Initialize} target network weights: \(\bar{\theta}_1 \gets \theta_1, \bar{\theta}_2 \gets \theta_2\)
\State \textbf{Initialize} empty replay buffer \(R\)

\For{each episode from \(1\) to \(N\)}
    \For{each timestamp \(t = 1\) to \(T\)}
        \State Reset exploration noise \( \epsilon \)
        \State Initialize state \((s_0^{(t)})\) = \(( v_n ^{(t)}, \, \Delta v_{n-1,n} ^{(t)}, \,  S_{n-1,n} ^{(t)}, \,   d_{tl} ^{(t)}, \, TL^{(t)})\) 
        \For{each timestamp \(t = 1\) to \(T\)}
        \State Select action: \( a^{(t)} = \pi_{\phi}(a^{(t)}|s^{(t)}) + \epsilon^{(t)} \)
        \State Execute action \( (a^{(t)}) \), observed reward \(( r^{(t)}) \), and next state \( (s^{(t+1)})\)
        \State Store transition tuple \( (s^{(t)}, a^{(t)}, r^{(t)}, s^{(t+1)}) \) in \( R \)
        \State Sample batch from \( R \)
        \State Update critic network:
        \State For each critic \(i\in \{1,2\}\)
         \(\theta_i \gets \theta_i - \lambda_Q \hat{\nabla}_{\theta_i} J_Q(\theta_i)\)
        \State where  \(\lambda_Q\) is the learning rate and \(J_Q(\theta)\) is the critic loss
        \State Update actor network:
        \(\phi \gets \phi - \lambda_{\pi} \hat{\nabla}_{\phi} J_{\pi}(\phi)\)
        \State Adjust temperature parameter:
       \(\alpha \gets \alpha - \lambda_{\alpha} \hat{\nabla}_{\alpha} J(\alpha)\)
        \State Update target network:  
        \State For each critic target \(i\in \{1,2\}\)
        \(\bar{\theta}_i \gets \tau \theta_i + (1-\tau) \bar{\theta}_i\)
        \EndFor
    \EndFor
\EndFor
\end{algorithmic}
\end{algorithm}

\begin{table}[h]
    \caption{Hyperparameters utilised in the SAC algorithm}
    \label{table: sac_table}
    \centering
    \renewcommand{\arraystretch}{1} 
    \begin{tabular*}{0.9\linewidth}{@{\extracolsep{\fill}} l l l }
        \toprule
        Symbol & Description & Value \\
        \midrule
        $LR_A$ & Actor learning rate (ADAM optimiser) & $0.001$ \\
        $LR_C$ & Critic learning rate (ADAM optimiser) & $0.002$  \\
        $\gamma$ & Discount factor for future reward calculation & $0.99$ \\
        $var_{a}$ & Initial variance of actions for exploration & $3$ \\
        $\lambda$ & Rate of variance reduction for exploration & $0.0005$ \\
        $B$ & Mini-batch size used for training & $512$ \\
        $\tau$ & Soft update rate for target networks & $0.001$ \\
        $R$ & Capacity of the replay memory buffer & $300000$ \\
        $N_l$ & Number of layers in actor and critic networks & $5$ \\
        $N_{hl}$ & Number of neurons in the hidden layers of the actor and critic networks & $128$ \\ 
        $\alpha$ & Initial temperature & $-2$ \\ 
        $T_{\alpha}$ & Temperature learning rate (ADAM optimiser) & $0.001$ \\ 
        $\bar{H}$ & Target entropy & $-2$ \\ 
        \bottomrule
    \end{tabular*}
\end{table}

\section{Data setup}
\label{sec: data and setup}
\begin{figure}[!ht]
    \centering
    \includegraphics[width = 0.95\linewidth]{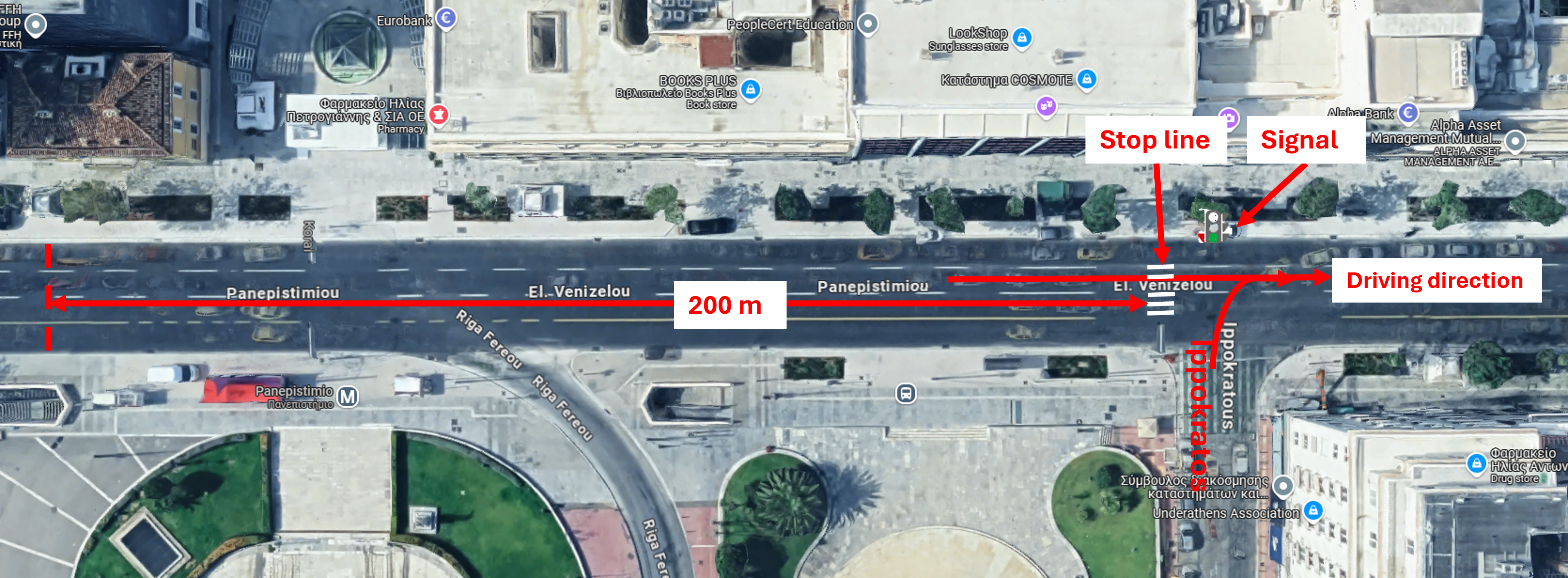}
    \caption{Geographical representation of the study site for the proposed RL framework}
    \label{fig: location}
\end{figure}

RL model training is based on the rewards received from different actions and therefore does not require human driving data as labels, unlike supervised learning. However, for vehicle manoeuvre learning, we need LV information. In previous RL-based vehicle movement learning studies, the LV trajectories have been obtained from the real-world \citep{zhou2019development, zhu2020safe, shi2022deep, yang2024eco}, simulation \citep{guo2021hybrid, qu2020jointly, wegener2021automated}, or both \citep{li2023modified, hart2024towards}. In our current study, we have utilised a combination of real-world pNEUMA dataset \citep{barmpounakis2020new} and simulated vehicle trajectories using the Ornstein–Uhlenbeck (OU) process \citep{ornstein1930theory}.
The pNEUMA dataset consists of high-resolution (25 Hz) trajectory data collected via a swarm of drones over a congested area in central Athens, Greece. Data was recorded for four weekdays, covering an area of around 1.3 $km^2$. For this study, we have utilised the dataset collected during the morning peak hours on November 1, 2018, at a T-intersection of Panepistimiou and Ippokratos street. Panepistimiou street is a one-way, three-lane road that experiences significant traffic congestion, particularly during peak hours. The distance from the stop line to the end of the signal influence area is approximately $200\ m$, as illustrated in Fig.~\ref{fig: location}. Although the pNEUMA dataset does not include the signal information, such as cycle length and phase timings. \cite{zhou2021queue} successfully extracted this information using a density-based non-parametric clustering algorithm (DBSCAN) to infer the signal timing based on the observed vehicle trajectories. \cite{zhou2021queue} found that the traffic signal at the T-intersection operated with a cycle length of 90 \textit{secs} and a phasing sequence of [16, 19, 45] secs. This means the green phase begins at 0  \textit{secs} and ends at 16 \textit{secs}, followed by a 3 \textit{secs} yellow phase from 16 to 19 \textit{secs}. The red phase starts at 19 \textit{secs} and continues until 45 \textit{secs}, after which the cycle repeats. 

A total of 158 pairs of car-following trajectories were extracted from the pNEUMA dataset, along with corresponding traffic light phase information. This pNEUMA dataset and inferred signal information have also been used in the previous study \cite{yang2024eco}. 
Additional LV trajectories were generated using the Ornstein–Uhlenbeck (OU) process \citep{ornstein1930theory}, which is discussed in the next section. These simulated LV trajectories help to provide RL models with a large number of diverse trajectories, which can help in robust model training.

\subsection*{Ornstein–Uhlenbeck (OU) process for leader vehicle trajectory}
\label{subsec: ou_process}

The OU process is a continuous-time stochastic process that describes how a system, such as vehicle speed, gradually returns to its normal state after random changes. Mathematically, it can be expressed as:
 \begin{equation}
          dv^{(t)} = \theta(\mu - v^{(t)})dt + \sigma dw^{(t)}
          \label{eq: Ou_contineous}
 \end{equation}
where $dv^{(t)}$ denotes stochastic change in speed at time $t$ and $\mu$ represents long term mean speed. $\theta$ denotes the rate at which the deviation of vehicle speed from the long-term average value $(\mu)$ over time. A higher rate of deviation makes the speed stabilise faster, while a lower rate provides slower adjustments. $\sigma$ represents the volatility, which controls the speed variation and $dw^{(t)}$ represents the Wiener process. All these parameters play an important role in generating accurate and dynamic vehicle trajectories. 
All parameters of the stochastic process are defined to approximate the kinematics of LV while ensuring realistic acceleration and deceleration behaviour. The acceleration/deceleration range of the LV is set to be $-2\ m/s^2$ to $2\ m/s^2$, which corresponds to a comfortable acceleration/deceleration $(a_{comf})$ \citep{treiber2000congested}. The parameters $\mu$ and $\theta$ depend on the speed limit of the road, which is $15\ m/s$ (i.e., $54\ km/h$). The long-term mean speed is estimated as $\mu = v_{des}/2 = 7.5 m/s$, while mean reversion rate is calculated as $\theta = a_{comf}/v_{des} =0.133 s^{-1}$. The volatility parameter $(\sigma)$, which determines the influence of random change in speed, is given by $\sigma = \sqrt{v_{des} a_{comf}/2}$.


To obtain a numerical solution, Eq.\eqref{eq: Ou_contineous} is discretised using the Euler–Maruyama method described in \cite{okhrin2022simulating}. This approach allows for changing the continuous-time OU process into its discrete-time form. The discretised version of the process is given by:
\begin{equation}
    v_{n-1}^{(t+1)} = v_{n-1}^{(t)} + \theta \left\{ \mu - v_{n-1}^{(t)} \right\} \Delta T + \sigma \Delta W
    \label{eq: OU_disc}
\end{equation}
with $\Delta W \sim N(0, \Delta T)$.

\begin{figure}[htb]
    \centering
    \includegraphics[width=1\linewidth]{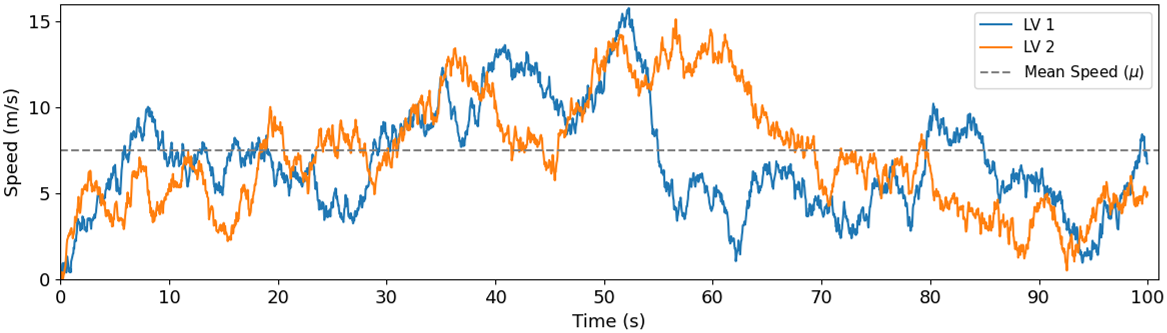}
    \caption{Sample of LV trajectories generated using a parametrised Ornstein–Uhlenbeck process for RL training}
    \label{fig: ou_trajectories}
\end{figure}

Fig.~\ref{fig: ou_trajectories} illustrates a sample of LV trajectories over a 100-secs duration, generated using parametrised OU processes. These LV trajectories exhibit a range of speed variations, including the intervals (e.g. 50 secs to 60 secs) when the speed reaches close to the desired value; on the other hand, the speed approaches zero between 90 secs to 95 secs. Such high variability in speed produces unpredictable LV behaviour, reflecting real-world driving conditions, where vehicles frequently accelerate and decelerate due to road and traffic constraints. Such fluctuations also lead to a diverse set of safety-critical events, including sudden stops and rapid accelerations. This results in enhancing the robustness of the RL agent when encountering unseen scenarios.

\section{Results and discussion}
\label{sec: results_discussion}
This section includes the training process of the RL agent to manoeuvre through the SI. The agent has been trained to take optimal acceleration/deceleration actions while maintaining safety, efficiency, and adherence to traffic regulations. 
After the training process is completed, the model has been compared with the real-world data for overall performance analysis using Empirical Cumulative Distribution Function (ECDF) plots. Further, individual sample trajectories are extracted and compared to show how the proposed model performs in different critical scenarios. 

\subsection{Training}
\label{subsec: training}
A combination of real-world (pNEUMA) and simulated LV trajectories (OU process) has been used to train the RL agent for vehicle movement at SI. The training dataset consists of 95 trajectories (60 \%) of the entire dataset from the pNEUMA dataset along with 200 simulated trajectories generated using the OU process \citep{ornstein1930theory}. 
The remaining 63 (40\%) of the dataset's real-world trajectories have been used for testing purposes. 
Throughout the training phase, the RL agent randomly selects a leader vehicle trajectory from the training dataset in every episode. 
The hyperparameters are manually tuned to optimise the performance across multiple trials. Although various automated techniques are available to optimise the hyperparameters \citep{espeholt2018impala, jomaa2019hyp, talaat2023rl}, manual tuning has been chosen for greater controllability and interoperability. 
Due to the stochastic nature of the reward function and environment, getting a truly optimal configuration is quite challenging. Therefore, expert judgment plays a crucial role in selecting hyperparameters. Previous RL-based CF studies have also performed manual tuning of hyperparameters \citep{zhou2019development, zhu2020safe, qu2020jointly, guo2021hybrid,  shi2023physics, yang2024eco}. However, in future, automated hyperparameter tuning can also be explored. The hyperparameters for the DDPG and SAC algorithms are listed in Table~\ref{table: ddpg_table} and Table~\ref{table: sac_table}.

\begin{figure}[htbp]
    \centering
    \includegraphics[width=0.7\linewidth]{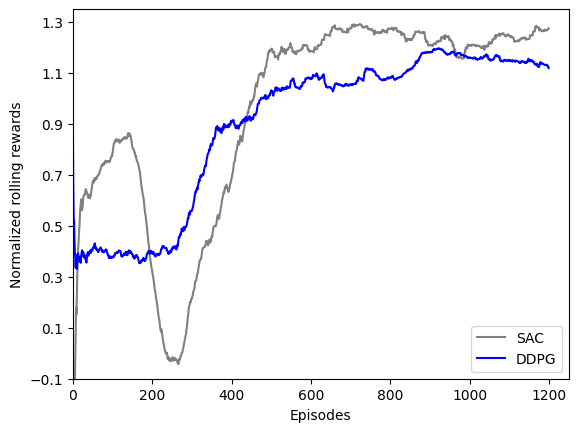}
    \caption{Normalised rolling rewards of the DDPG and SAC models during training}
    \label{fig: rolling_score}
\end{figure}

Fig.~\ref{fig: rolling_score} shows the training performance of the two algorithms, DDPG and SAC with the proposed reward function. The y-axis shows the normalised rolling average reward, calculated with min-max normalisation on a sliding window of 100 steps, while the x-axis shows the training episode.
As illustrated in Fig.~\ref{fig: rolling_score}, the DDPG model shows a noticeable increase in the rolling score during the first 400 episodes and then stabilises. On the other hand, in the SAC model, the rolling score declined sharply between 200 to 300 episodes, followed by a sharp increase to 600 episodes and then stabilised.
 The maximum rolling achieved by the SAC model while training is slightly higher than DDPG model. The SAC model exhibits lesser fluctuation than DDPG after stabilisation around 800 episodes.
 We stopped the training procedure after the 1200 episodes when the training process stabilised. The trained model has been used to generate the test results, which will be discussed in the next section.
 
 
\subsection{Test results}
\label{subsec: test_results}
Our primary objective is to test the trained model based on the defined criteria, which includes safety, efficiency, comfort, compliance with traffic signals, and desired speed across various driving scenarios. To comprehensively assess the robustness of the model, we focus on two distinct test scenarios.
First, we compare the model's overall performance with real trajectory data by analysing the ECDF of TTC, DH, and jerk. These metrics provide valuable insights into the overall safety, efficiency, and comfort of the driving behaviour exhibited by the trained model.
Secondly, we evaluate the individual trajectory-wise performance by checking a few sample trajectories of the RL model under different critical conditions. Details of the test data setup and results obtained are discussed next. 



\subsubsection{Overall trajectory performance results}
\label{subsec: overall_traj_performance}
The objective here is to compare the overall performance of the trained RL agent trajectories with the real-world FV test trajectories. The LV trajectories in both cases were extracted from the pNEUMA test dataset, comprising real data trajectories. This comparison has been done using ECDF plots, similar to \cite{zhou2019development, zhu2020safe}. The comparison has been done on the basis of three metrics: (i) Safety based on TTC, (ii) Efficiency based on DH, and (iii) Comfort based on jerk. Each of these metric comparisons are discussed next. Also, no collisions were recorded by the RL agent for all real data LV trajectories. 

\subsection*{Safety}
Safety is evaluated based on TTC while following LV as defined in Eq.~\eqref{eq: f_TTC}. Fig.~\ref{fig: ECD}(a) shows the comparison of the cumulative distribution of the TTC values from the DDPG model, SAC model, and human FV trajectories data, denoted as real data. For better interpretability, TTC values have been considered within the range of 0 to 50 $secs$. Ideally, the CDF obtained from the models should be shifted to the right compared to real data, which would mean that the trained agents are able to keep a larger TTC, ensuring more safety.  Fig.~\ref{fig: ECD}(a) illustrates that the CDF of both the DDPG and SAC models is closely aligned with real data and shifts further right for TTC greater than 16 secs. Overall, it can be stated that both the DDPG and SAC models are able to maintain a similar TTC as human driver vehicles.

\subsection*{Efficiency}
The efficiency reward encourages the agent to keep the DH close to the desired DH, given by Eq.~\eqref{eq: f_eff}. Fig.~\ref{fig: ECD}(b) shows the comparison of the ECDF of the DH obtained from the DDPG model, SAC model, and human FV trajectories. It can be observed that the ECDF of both DDPG and SAC models shifted towards the left side compared to the real FV trajectories, indicating that the models maintain a lower DH than that typically maintained by human drivers. However, it can be noted that the DDPG and SAC models are able to keep headway close to the corresponding desired headway, shown using the dashed line in Fig.~\ref{fig: ECD}(b). Therefore, it can be concluded that both models, DDPG and SAC, keep TTC close to real-world trajectories while maintaining lower DH values, demonstrating an effective balance between safety and efficiency.


\subsection*{Comfort}
\begin{figure}[htbp]
    \centering
    \includegraphics[width=1\linewidth]{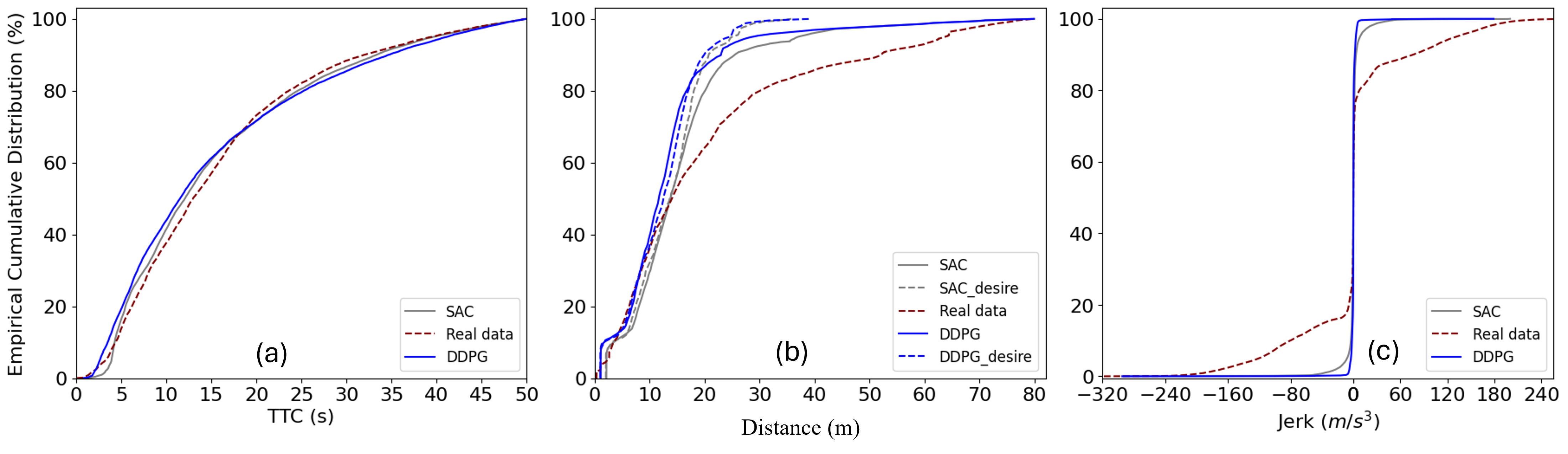}
    \caption{Empirical CDFs of: (a) TTC (b) space, and (c) jerk for RL agents based on leader trajectory}
    \label{fig: ECD}
\end{figure}

The final overall performance measure considered in this section is driving comfort, which is measured in terms of a jerk, given in Eq.~\eqref{eq: f_jerk} during manoeuvring at the intersection. 
 Fig.~\ref{fig: ECD}(c) shows the comparison of the cumulative distribution of the jerk obtained from the DDPG model, SAC model, and human FV trajectories. It can be seen that the ECDF of both the DDPG and SAC models maintains a jerk value close to zero, indicating comfortable driving compared to the real FV trajectories. Although the jerk data extracted from real-world trajectories indicate that human driving leads to higher jerks, however, it can be noted that the jerk values are obtained from the vehicle coordinate values extracted from the drone data. Since the position data inherently contains some noise. Therefore, noise in the vehicle position data extraction can lead to higher values of jerk (third derivative of x-position). In the real world, the actual jerk might have been lower than the jerk values recorded in the data. Nonetheless, the trained DDPG and SAC models were able to maintain significantly lower values of jerk, thereby indicating a comfortable ride. 
 The average jerk for both DDPG and SAC models was (0.63 $m/s^3$) followed by SAC (3.25 $m/s^3$).  Additionally, we also calculated the percentage of the time the jerk value was lower than the comfortable jerk of 1.5 $m/s^3$ \citep{treiber2000congested}. The DDPG model has (9.52$\%$) followed by SAC (30.16$\%$) and real driver (31.75$\%$). Therefore, it can be concluded that the DDPG model performed best in terms of all three parameters: safety, efficiency, and comfort.
 

\subsubsection{Sample trajectory performance results}
While ECDF plots provide an overall comparison of the proposed RL models, it is also important to test individual trajectories and check how the models perform on an individual scale. The evaluation includes three key aspects: (i) Comparison with each real-world LV trajectory approaching the intersection during different phases of a traffic signal, (ii) Testing the models in critical simulated car-following scenarios, and (iii) Assessing compliance with traffic signals. Each of these comparisons and tests is discussed in detail in the following subsections.

\subsection*{Real trajectory performance}
\begin{figure}[h]
    \centering
    \includegraphics[width=1\linewidth]{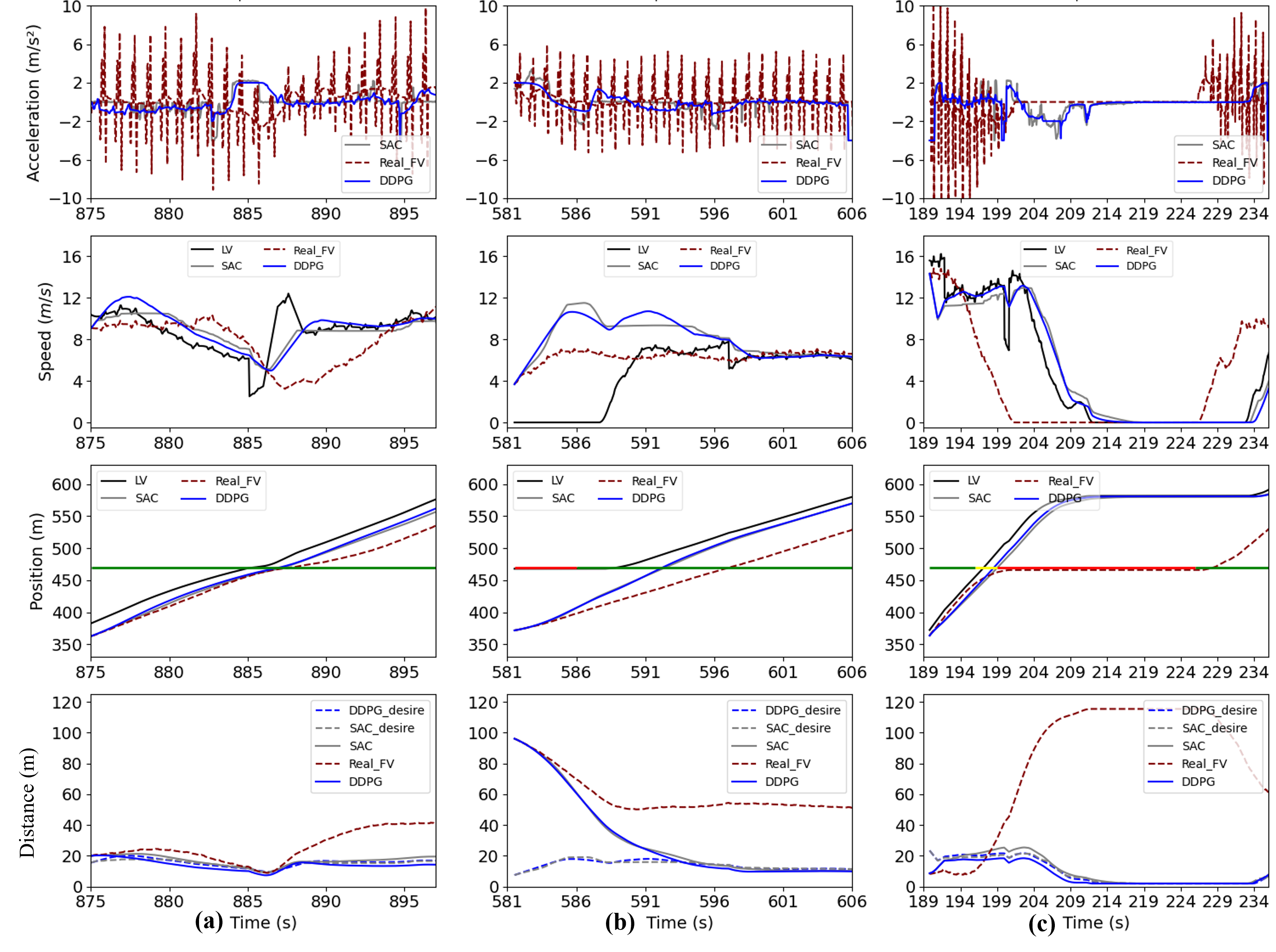}
    \caption {Response to real LV trajectory approaching the intersection during: (a) middle of green, (b) start of green, (c) towards the end of green and amber}
    \label{fig: real_cf}
\end{figure}

To provide more insight into the behaviour of the trained RL models while approaching traffic intersections, as shown in Fig.~\ref{fig: real_cf}.
Fig.~\ref{fig: real_cf}(a) (first column) shows the scenario where the LV and the FV (agent) approaches the intersection during the middle of the green phase, Fig.~\ref{fig: real_cf}(b) (2nd column) represents the situation where the agent arrives at the start of the green phase while the LV was stationary at the stop line during red light, and finally, Fig.~\ref{fig: real_cf}(c) (3rd column) depicts the scenario where the agent approaches the intersection toward the end of the green phase or during the amber phase. For each scenario, we provide the plots for acceleration, speed, longitudinal position, and DH variation with respect to time. The black, blue, and gray solid lines denote the LV, trained DDPG FV, and trained SAC FV, respectively. The red dotted lines denote the response of the human-driven FV, denoted as Real FV. For the DH plots (last row of Fig.~\ref{fig: real_cf}), we also show the variation of the desired DH defined in Eq.~\eqref{eq: space_headway} using a blue dotted line for trained DDPG FV and a gray dotted line for trained SAC FV.
It can be seen from Fig.~\ref{fig: real_cf} that the RL agents consistently adapt their speed profile based on the LV trajectories and follow signal rules to ensure compliance with traffic signals. Fig.~\ref{fig: real_cf}(a) shows that when the LV decelerates, the RL agent also decelerates (but with a smoother deceleration profile), followed by reaching a speed close to that of the LV. This shows that the RL agent is capable of demonstrating regular car-following behaviour.
A few other interesting observations can also be made from Fig.~\ref{fig: real_cf}. The trained RL agent (FV) tries to maintain zero relative speed in all three cases when the situation permits (except for large acceleration/deceleration by LV). This phenomenon, known as ``\textit{local stability}", where the FV stabilises at a safe DH and zero relative speed after initial perturbation of the LV actions, is one of the important car-behaviour characteristics. Although the RL model was not explicitly trained to follow this behaviour in terms of the rewards function, however, the agent learned this behaviour through training. It can be further noted that in Fig.~\ref{fig: real_cf}(a) when the LV decelerates at around 885 secs and then accelerates, the RL agent also follows the same behaviour, but comes back to the previous safe DH of around 20 m, unlike the Real FV (human-driven) maintains a higher DH (around 40 secs) despite the same speed of LV ($\sim$ 10 secs) before and after the perturbation. Therefore, although the RL agent maintains a lower DH than the real FV. However, the DH is close to the safe DH provided to the agent. This finding strengthens our proposed approach, which does not imitate human driving behaviour through data-driven deep learning, but rather leverages RL to optimise vehicle control based on a carefully designed reward function. Overall, this highlights that the RL model can successfully maintain traffic signal rules, follow realistic car-following behaviour with smoother acceleration and deceleration profiles, and maintain stable, safe DH with the LV.

 

\subsection*{Performance under critical driving situations}
\begin{figure}[h]
    \centering
    \includegraphics[width=1\linewidth]{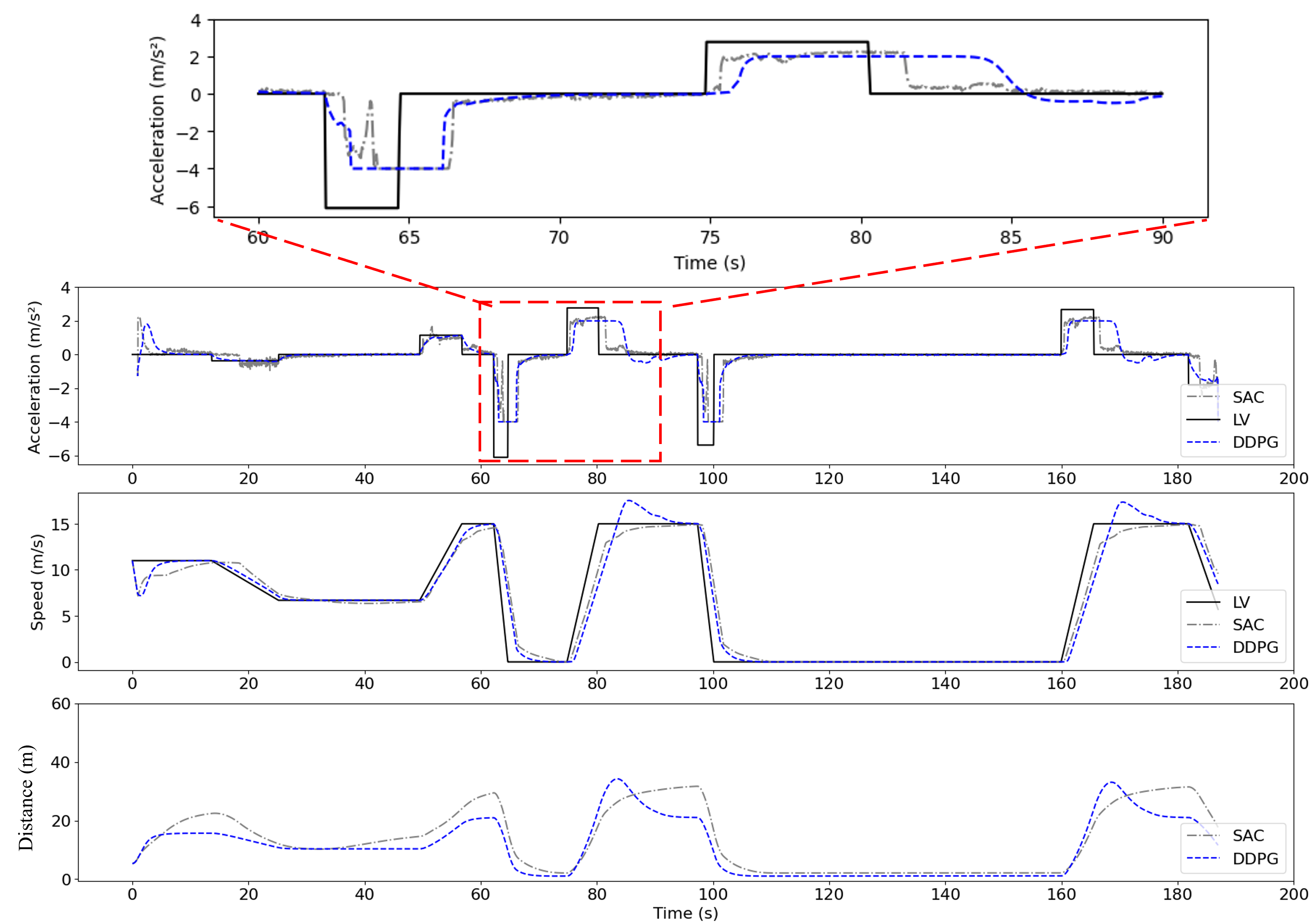}
    \caption{Response of RL models in extreme driving scenarios with simulated LV trajectories}
    \label{fig: CF}
\end{figure}

Testing the RL agent using with real-world trajectory dataset allows us to compare the model performance with human drivers. However, the limited sample size of the real-world trajectories makes it difficult to test the model for critical scenarios such as sudden deceleration. Since one of the major motivations in building autonomous vehicles is to handle all kinds of critical scenarios. Therefore, we created such a scenario that can test the robustness of the RL model against safety-critical events.
We simulated the LV trajectory, as shown by the black solid line in the acceleration and speed plots in Fig.~\ref{fig: CF}. The initial speed of the LV was kept at $11 \ m/s$, followed by a constant deceleration of 0.5 $m/s^2$ from 18 secs to 25 secs and acceleration of 1 $ m/s^2$ between 50 secs to 55 secs.
Further, to assess the agent’s response to sudden deceleration, we introduced a sharp deceleration of $-6 \,m/s^2$ and  $-5 \,m/s^2$ at $t=62\,secs$ and $t=98\,secs$ respectively, resulting in the vehicle becoming stationary at both instants. It can be noted that the deceleration rate of LV was kept greater than the maximum allowed deceleration of FV ($-4 \,m/s^2$). This was done to intentionally check whether the FV can successfully avoid collision even in this extreme situation.

Both DDPG and SAC models successfully handled the sudden deceleration by applying smooth braking within the acceleration bound of $a_{min} = -4\,m/s^2$ ensuring a minimum DH of approximately $s_0=2\,m$ from the leader vehicle within the time interval of $[0,200\, secs]$. As expected, the maximum acceleration and deceleration of FV were bounded to $+2 \,m/s^2$ and $-4 \,m/s^2$ respectively, which ensures the asymmetric behaviour of the agent towards acceleration and deceleration events of the LV, incorporated through the acceleration reward function $(f_{Acc}^{(t)})$ give in Eq. ~\eqref{eq: f_action}.
As illustrated in the zoomed-in view of the acceleration plot at the top of Fig.~\ref{fig: CF}, both RL models exhibit smooth and slightly delayed deceleration and acceleration responses, which contribute to a comfortable driving experience, even when the leader vehicle applies hard braking. 
However, the SAC model shows minor fluctuations in action values while approaching the leader vehicle during sudden braking events, as illustrated in Fig.~\ref{fig: CF} (zoomed-in view). In contrast, the DDPG model maintains a more stable and smooth acceleration-deceleration profile. Furthermore, the SAC model tends to maintain a slightly larger minimum following distance $(s_0=2\,m)$ compared to the DDPG model, which could contribute to a more conservative and cautious driving behaviour.

\subsection*{Traffic signal scenarios}
Along with the simulated scenario for safety-critical events, we also tested the RL model performance for traffic signal compliance under various simulated scenarios using OU process \citep{ornstein1930theory} as described in Section ~\ref{subsec: ou_process}. These scenarios include different combinations of start time, initial speed, and acceleration of the FV. Here, start time refers to the time when the FV enters the intersection boundary i.e., 350 m where 0 secs refers to the start of the green signal. To recall, the signal phase timing used in this study is green signal 0-16 secs, amber 16-19 secs, and red 19-46 secs. 
The start time of FV for simulation was varied from 0 to 45 secs, with different increments to ensure a robust evaluation of the model's response to the amber light. Initially, the start time increased in increments of 5 secs for up to 15 secs [i.e., 0, 5, and 15 secs], followed by finer increments of 1 secs from 15 to 20 secs to capture how the FV behaves while approaching the amber light. After 20 secs, the increments reverted back to 5 secs to maintain a broad range of test conditions. The initial speed of FV was taken from 0 $m/s$ to 14 $m/s$ with increments of 2 $m/s$, ensuring a diverse set of speeds. Additionally, the starting acceleration was randomly selected between 0 to 1 $m/s^2$ to introduce variability in vehicle dynamics and assess the model's adaptability under different conditions. In all these scenarios, the LV position started after the signal stop line to ensure that the FV (RL agent) makes driving decisions based on the traffic light only and does not get influenced by the LV position. The trajectory of the LV was generated using the OU process. In total, 128 simulated scenarios were created using various combinations of start time, initial speed and acceleration. 

\begin{figure}[h]
    \centering
    \includegraphics[width=1\linewidth]{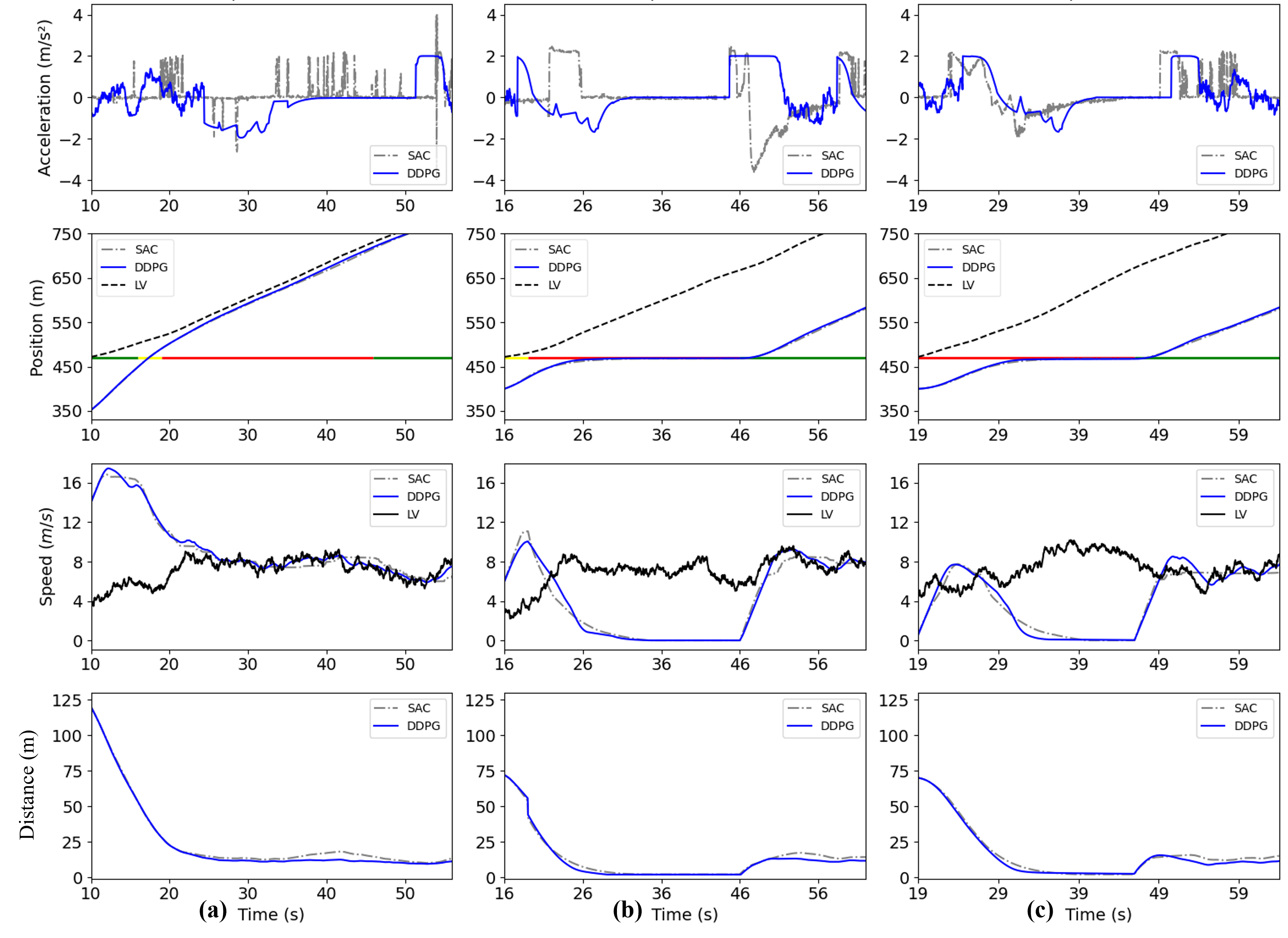}
    \caption{Performance evaluation of RL agents at the traffic signals during: (a) amber (crossing during  amber), (b) amber (stopping during amber), (c) red light (stopping during red)}
    \label{fig: signal_test}
\end{figure}

In all these simulated scenarios, both DDPG and SAC-trained models successfully complied with the traffic signal rules and didn't violate any red signal. Fig.~\ref{fig: signal_test} shows three representative trajectories to examine how models respond in different situations. These scenarios demonstrate how RL agents make decisions when approaching intersections under varying signal conditions.
Fig.~\ref{fig: signal_test}(a) represents the situation when FV approaches intersections during the amber light phase and chooses to cross the signal before turns red. This highlights the agent's ability to evaluate whether it can safely pass through the intersection while adhering to traffic regulations.
On the other hand, Fig.~\ref{fig: signal_test}(b) illustrates the situation when FV approaches the intersection during amber light and decides to stop instead of proceeding through the intersection. This scenario evaluates the agent's ability to anticipate the transition from amber to red and make a safe stopping decision. Although the FV decided to accelerate for a very brief period, indicated by the sharp increase in acceleration, it realises that it is not possible to cross the intersection with the amber light and hence decelerates smoothly to come to a stop. Once stopped, the vehicle remains stationary throughout the entire red-light phase, ensuring compliance with traffic rules. It can be noted that the DDPG model exhibited a smoother acceleration/deceleration profile compared to the SAC model.

Finally, Fig.~\ref{fig: signal_test}(c) represents another scenario where the agent approaches the intersection while the traffic light is red. The agent correctly identifies the red light and stops, ensuring compliance with traffic rules and avoiding potential violations. The acceleration plot indicates a significant acceleration during the initial phase of the red light and then deceleration, suggesting that the vehicle appropriately responds to the red traffic signal. The speed remains at zero throughout the red light phase. Once the signal turns green, both models accelerate with maximum possible acceleration, demonstrating a prompt response to the change in traffic conditions. It can be noted that the maximum acceleration in this study has been taken as $+2 m/s^2$, which is the comfortable acceleration rate used in previous studies. Therefore, it can be concluded that both the DDPG and SAC models complied with all traffic signal scenarios, showing the robustness of the models, and the DDPG model showed smoother action profiles compared to the SAC model. This aligns with the results shown in the Section where the DDPG model performed better compared to the SAC model.

\section{Conclusion}
\label{sec: conclusion}
This study presents an RL-based longitudinal vehicle control strategy at signalised intersections. A comprehensive reward function has been proposed, taking into consideration (i) safety (TTC), (ii) efficiency (distance headway), (iii) traffic signal compliance (with particular emphasis on decision making during amber light), (iv) speed compliance, (v) assymetric acceleration/deceleration response, and (vi) comfort (jerk). Two widely used RL algorithms, DDPG and SAC, have been utilised to develop an adaptive and robust vehicle control strategy using the proposed reward function. 
The training scenario has been built by integrating real-world naturalistic driving data from the pNEUMA dataset with simulated leader vehicle trajectories based on the OU process. The randomness of the OU process allows the simulation of critical driving events that are typically under-represented within real-world datasets, thereby strengthening the robustness of RL agents.

We have tested it under different unseen situations, including both real-world and simulated trajectories of the leader vehicle, to assess the robustness and generalisation of the proposed model. 
The model performance has been measured using important metrics such as compliance with traffic signals, stability of acceleration and deceleration, safety in response to sudden breaks, efficiency to enhance traffic flow, and ensuring a comfortable driving experience. The results show that the RL-based vehicle control system effectively maintains a safe following distance, responds well to traffic signals, and provides smooth acceleration and deceleration. Although both DDPG and SAC models successfully handled these test scenarios, the trained DDPG showed slightly better results than the SAC model in terms of a smoother acceleration profile (resulting in lower jerk) for all real-world and safety-critical scenarios.
The results confirm that RL-based vehicle control strategy with careful selection of the reward function and incorporating both real-world and simulated data can help the model succeed in learning desired driving behaviour and also maintain robustness under extreme circumstances. 

This work can be extended in the future by considering fuel efficiency in the reward function to develop eco-driving strategies while manoeuvring traffic intersections. Further, this study used a limited set of real-world trajectories for training and testing purposes. To handle this, we used simulated trajectories to create a larger and diverse dataset. However, in future, the study dataset can be expanded with large-scale real-world trajectories with diverse locations and driving behaviour. This can help to assess the model's performance and compare it with real-world data. Finally, the trained RL models can be applied in real-world settings, even if it is at a limited scale, to analyse the transferability and adaptability in realistic traffic conditions.


\section*{CRediT authorship contribution statement}
\textbf{Pankaj Kumar:} Conceptualisation, Formal analysis, Methodology, Investigation, Writing. \textbf{Aditya Mishra:} Formal analysis, Methodology, Investigation, Writing.
\textbf{Pranamesh Chakraborty:} Conceptualisation, Methodology, Investigation, Supervision, Writing.  
\textbf{Subrahmanya Swamy Peruru:} Conceptualisation, Methodology, Investigation, Supervision, Writing.  

\section*{Declaration of competing interest}
The authors confirm that they have no financial or personal conflicts of interest that could have influenced the findings of this study.

\bibliography{bibliography}
\end{document}